\documentclass{article}
\usepackage{amsmath}
\UseRawInputEncoding
\usepackage[utf8]{inputenc}

\usepackage{arxiv}
\usepackage{float}
\usepackage{tikz}
\usetikzlibrary{arrows.meta, positioning}

\usepackage[utf8]{inputenc} 
\usepackage[T1]{fontenc}    
\usepackage{hyperref}       
\usepackage{url}            
\usepackage{booktabs}       
\usepackage{amsfonts}       
\usepackage{nicefrac}       
\usepackage{microtype}      
\usepackage{multirow}       
\usepackage{lipsum}
\usepackage{graphicx}
\usepackage{bm}
\usepackage{subcaption}
\usepackage[numbers]{natbib}
\usepackage{tikz}
\usetikzlibrary{positioning, fit, shapes.geometric}
\usepackage{graphicx}
\usepackage[hypcap=false]{caption} %
\graphicspath{ {./images/} }

\title{Post-Training Quantization of \\Generative and Discriminative LSTM Text Classifiers: \\A Study of Calibration, Class Balance, and Robustness}

\author{
Md Mushfiqur Rahaman\\School of Mathematical and Data Sciences\\ West Virginia University\\Morgantown, WV, 26505, USA\\\texttt{mr00131@mix.wvu.edu} \\
\And
Elliot Chang\\School of Mathematical and Data Sciences\\ West Virginia University\\Morgantown, WV, 26505, USA\\ \texttt{sc00048@mix.wvu.edu} \\
\And
Tasmiah Haque\\Department of Industrial and Management Systems Engineering\\ West Virginia University\\Morgantown, WV, 26505, USA\\ \texttt{th00027@mix.wvu.edu} \\
\And
Srinjoy Das\thanks{Corresponding author.}\\
School of Mathematical and Data Sciences\\
Department of Industrial and Management Systems Engineering\\
West Virginia University\\
Morgantown, WV, 26505, USA \\
\texttt{srinjoy.das@mail.wvu.edu} \\
}

\begin{document}
\maketitle
\begin{abstract}
Text classification plays a pivotal role in edge computing applications like industrial monitoring, health diagnostics, and smart assistants, where low latency and high accuracy are both key requirements. Generative classifiers, in particular, have been shown to exhibit robustness to out-of-distribution and noisy data, which is an extremely critical consideration for deployment in such real-time edge environments. However, deploying such models on edge devices faces computational and memory constraints. Post Training Quantization (PTQ) reduces model size and compute costs without retraining, making it ideal for edge deployment. In this work, we present a comprehensive comparative study of generative and discriminative Long Short Term Memory (LSTM)-based text classification models with PTQ using the Brevitas quantization library. We evaluate both types of classifier models across multiple bitwidths and assess their robustness under regular and noisy input conditions. We find that while discriminative classifiers remain robust, generative ones are more sensitive to bitwidth, calibration data used during PTQ, and input noise during quantized inference. We study the influence of class imbalance in calibration data for both types of classifiers, comparing scenarios with evenly and unevenly distributed class samples including their effect on weight adjustments and activation profiles during PTQ. Using test statistics derived from nonparametric hypothesis testing, we identify that using class imbalanced data during calibration introduces insufficient weight adaptation at lower bitwidths for generative LSTM classifiers, thereby leading to degraded performance. This study underscores the role of calibration data in PTQ and when generative classifiers succeed or fail under noise, aiding deployment in edge environments.
\end{abstract}


\section{Introduction}

Text classification remains a foundational task in natural language processing (NLP), with wide-ranging applications such as sentiment analysis, topic detection, and spam filtering\cite{zhang2015}. Modern deep learning models---particularly those based on Long Short Term Memory (LSTM) \cite{hochreiter1997long}, Transformers \cite{vaswani2017attention}, and other architectures which can process sequential data---have achieved state-of-the-art results on standard text classification datasets \cite{howard2020,zhang2015_AgNews_DBPedia}. However, these models are often too large and computationally intensive for real-time inference on small form-factor edge devices, such as smartphones, microcontrollers, or IoT nodes, which necessitates the requirement of performing model reduction without degrading performance.

Model quantization, which is the process of reducing the precision of weights and/or activations of a model, offers a practical solution to this challenge. Lower bitwidth representations (e.g., 8-bit, 4-bit, or 3-bit) can significantly reduce memory usage and improve inference speed, while often maintaining similar accuracy to full-precision models \cite{liu2024spinquant, zhang2025qronos}. A particularly appealing approach is Post-Training Quantization (PTQ), which does not require model retraining or full access to labeled training data  \cite{zhang2025qronos, hubara2021accurate}. Tools like Brevitas \cite{brevitas} enable flexible, hardware-aware PTQ workflows, making them suitable for real-world edge computing scenarios.

While a significant amount of research has focused on quantizing discriminative classifiers \cite{wang2019haq, zafrir2019q8bert}, which are models that are trained to directly map inputs to labels, generative classifiers which model class-conditional distributions  remain underexplored in the quantization literature. This gap is notable given that generative classifiers have demonstrated superior robustness to input noise and out-of-distribution (OOD) data \cite{ ktena2024generative, lee2019robust, nguyen2015deep, ng2001discriminative, bishop2006pattern}, making them increasingly attractive for safety-critical and privacy-sensitive deployments .

A key challenge in deploying quantized models is that PTQ is often performed using passively collected, unlabeled data, especially in edge or privacy-restricted environments. This data is frequently class-imbalanced, which can lead to biased activation statistics and performance degradation, particularly in class-conditional models. Despite its practical relevance, the impact of data imbalance during PTQ remains underexamined in the literature.

In this work, we present a comprehensive comparison of generative and discriminative LSTM-based text classifiers under low bitwidth PTQ using Brevitas. We study how model performance changes across different bit-widths (from 8-bit down to 3-bit) and how class distribution of the data used for PTQ affects accuracy, noise robustness, and internal weight/activation distributions. We use Kolmogorov–Smirnov (KS) tests \cite{kolmogorov1933empirical, smirnov1948table} to analyze and quantify shifts in the activation distributions and track the root causes of performance degradation during PTQ, particularly in generative LSTM classifiers. Finally, we conduct input noise robustness tests to simulate real-world data perturbations and compare how the two types of LSTM classifiers quantized using PTQ respond under non-ideal conditions.

Our results show that generative classifiers are significantly more sensitive to both class imbalance of data used during PTQ and input noise, especially at lower bit-widths and offers practical insights for deploying such classifiers on real-world, resource constrained edge computing platforms. Our findings are novel in systematically analyzing post-training quantization of generative classifiers, highlighting the critical role of calibration data composition and class balance, which are aspects often overlooked in conventional quantization studies focused on discriminative models.

\section{Related Works}

This section reviews key developments in PTQ, including its application to discriminative and generative models, and the emerging role of data selection in achieving robust quantized performance. While significant progress has been made in quantizing discriminative models across vision and language tasks, generative models remain underexplored in this context.

\subsection{Post-Training Quantization (PTQ)}

Post-Training Quantization (PTQ) has been extensively explored as a model compression technique that enables efficient inference by converting full-precision models into lower bitwidth representations without requiring retraining. This approach is especially attractive for deployment on resource-constrained edge devices due to its simplicity and minimal reliance on training data \cite{nagel2019data, nagel2020up}. PTQ is now a standard feature in modern deployment toolchains such as PyTorch\cite{paszke2019pytorch}, 
particularly for feedforward networks, Convolutional Neural Networks (CNNs) \cite{xu2022easyquant}, and Transformer-based models like BERT and GPT \cite{yao2022zeroquant, li2023fp8, hu2022empirical, bondarenko2021understanding}. Most PTQ research to date has focused on discriminative classifiers which evaluate the impact of quantization granularity, clipping techniques, and calibration strategies on classification accuracy. Despite emerging evidence that generative classifiers offer greater robustness to noise and out-of-distribution inputs \cite{ ktena2024generative, lee2019robust, nguyen2015deep, ng2001discriminative, bishop2006pattern}, very few studies have assessed their behavior under quantization\cite{ruan2021generative}.

\subsection{Data Selection for PTQ}

Recent work by Williams and Aletras \cite{li2023impact} presents a systematic investigation into how the selection of PTQ datasets influences the model accuracies in large language models (LLMs). Their findings reveal that even minor variations in the source or composition of unlabeled PTQ data can lead to substantial fluctuations in downstream accuracy, as seen on open source LLM families such as LLaMA, Vicuna and OPT \cite{touvron2023llama, chiang2023vicuna, zhang2022opt}. The study identifies critical PTQ data issues, namely distribution shift, dataset overlap, and small size which can severely compromise PTQ outcomes.

Similarly, Zhang et al. \cite{zhang2025selectq} emphasize the importance of careful PTQ data selection in maximizing performance. They identify three key issues contributing to quantization degradation: random selection of data samples, distribution mismatch in activations and outlier sensitivity, which leads to inaccurate determination of activation ranges. These shortcomings intensify quantization noise and diminish model accuracy, especially under aggressive low-bit quantization regimes. To mitigate these effects, the authors propose SelectQ, a novel method that leverages activation-aware dynamic clustering to identify more representative data samples and ensure better alignment with the model’s activation distribution. To the best of our knowledge, no comparable studies have examined how PTQ data selection impacts generative classifiers, leaving an important gap in the literature.

\section{Discriminative and Generative LSTM Classifiers}

This section describes the architecture and training methodology of discriminative and generative classifiers, specifically based on Long Short-Term Memory (LSTM) networks. Discriminative classifiers directly optimize decision boundaries for classification, whereas generative classifiers model class-conditional distributions which are then used to perform classification using Bayes rule \cite{hastie2009elements}. Understanding their structural differences in case of LSTMs is essential for interpreting their behavior under quantization.

Before textual data can be processed by a neural classifier, it must first be converted from raw text into a numerical form. This transformation typically begins with \textit{tokenization}, a process that segments a sentence or document into discrete units called tokens which are usually formed from words or subwords. Each token is then assigned a unique index from a predefined vocabulary. This index-based representation allows models to operate on symbolic input efficiently, but it lacks semantic structure. To overcome this limitation, each token index is further mapped to a continuous vector representation, a process known as \textit{embedding}. Several standard techniques have been proposed to encode semantic relationships among words. Popular approaches include word2vec \cite{mikolov2013efficient}, which learns word representations based on context prediction; GloVe (Global Vectors for Word Representation) \cite{pennington2014glove}, which incorporates global word co-occurrence statistics; and modern NLP libraries like SpaCy\cite{honnibal2020spacy}, which provide pretrained embeddings and efficient tokenization pipelines. 

\subsection{Discriminative LSTM Classifier Architecture}

The discriminative text classification model used in this work is based on the Long Short-Term Memory (LSTM) architecture, which has been widely adopted for modeling sequential data due to its ability to capture long-range dependencies \cite{hochreiter1997long}. The model consists of an embedding layer, an LSTM encoder, and a fully connected output layer that directly maps a sequence of tokens to a class label.

Let the sequence of tokens generated from the input text be denoted as \( x = (x_1, x_2, \ldots, x_T) \), where each \( x_t \in \mathbb{Z}^+ \), $t \in \{1,2, \ldots, T\}$ represents the index of the \( t \)-th token in a vocabulary of size \( V \), and \( T \) is the length of the input sequence. Each token \( x_t \) is mapped to a dense vector \( \mathbf{e}_t \in \mathbb{R}^d \) of dimension \( d \) using an embedding matrix \( E \in \mathbb{R}^{V \times d} \):
\begin{equation}
    \mathbf{e}_t = E[x_t], \quad t = 1, \ldots, T
\end{equation}
This results in an embedded input sequence \( \mathbf{E} = (\mathbf{e}_1, \ldots, \mathbf{e}_T) \in \mathbb{R}^{T \times d} \), where \( \mathbf{e}_t \) is the embedding vector corresponding to token \( x_t \). The purpose of using embeddings is to convert discrete token indices into continuous vector representations that can be processed by the neural network. Instead of using high-dimensional one-hot encodings, the model learns a dense, low-dimensional embedding for each token during training. These embeddings capture useful patterns and relationships between tokens based on the classification objective, allowing the model to generalize better across similar input sequences \cite{jurafsky2023speech}.

\noindent The embedded sequence is then processed by an $L$ layer LSTM network over $T$ timesteps as below\cite{hochreiter1997long}:

$$
\forall t = 1,\ldots,T, \ \forall l = 1,\ldots, L: {\bf s}_t^{(l)} = F^{(l)}({\bf s}_{t-1}^{(l)}, {\bf u}_{t}^{(l)})
$$

\noindent where
\[
{\bf u}_t^{(l)} =
\begin{cases}
{\bf e}_t & \text{if } l = 1 \\
{\bf h}_t^{(l-1)} & \text{if } l > 1
\end{cases}
\]

\noindent and for a given layer $l$, the function for the LSTM computations at layer $l$ is denoted by $F^{(l)}$ and the combined state ${\bf s}_{t}^{(l)}$ is composed of the hidden state ${\bf h}_t^{(l)}$ and cell state ${\bf c}_t^{(l)}$ of the LSTM as below:
$$
{\bf s}_{t}^{(l)} = ({\bf h}_{t}^{(l)}, {\bf c}_{t}^{(l)})
$$

\noindent We use the hidden state of the last layer at the final time step as the internal representation of the input sequence:
\begin{equation}
    \mathbf{h}_{\text{final}} = \mathbf{h}_T^{(L)}
\end{equation}

\noindent This internal representation is passed through a fully connected linear output layer with weight matrix \( W \in \mathbb{R}^{C \times h} \), where \( C \) is the number of target classes and $h$ is the dimension of ${\bf h}_{\text final}$. The class scores or logits are computed as:
\begin{equation}
    \mathbf{z} = W \cdot \mathbf{h}_{\text{final}}
\end{equation}
where \( \mathbf{z} \in \mathbb{R}^C \) contains one score \( z_k \) for each class \( k \in \{1, \ldots, C\} \). The final class probabilities \( \hat{y}_k \) are obtained using the softmax function applied to the logits:
\begin{equation}
    \hat{y}_k = \frac{\exp(z_k)}{\sum_{j=1}^C \exp(z_j)}, \quad k = 1, \ldots, C
\end{equation}
where \( \hat{y}_k \in [0, 1] \) represents the predicted probability of the input sequence belonging to class \( k \), and the probabilities sum to 1 across all \( C \) classes.\\

The model is trained using the cross-entropy loss function, which compares the predicted probability \( \hat{y} \) with the one-hot encoded ground truth label $y$ for a given sample as below:

\begin{equation}
    \mathcal{L}_{\text{CE}} = -\sum_{k=1}^{K} y_{k} \log \hat y_{k}
\end{equation}
Here \( \mathcal{L}_{\text{CE}} \) denotes the cross entropy loss for one sample. 

\subsection{Generative LSTM Classifier Architecture}

Apart from the discriminative model, this work also explores a generative classification approach, where the model learns to generate an input sequence conditioned on a given class label \cite{yogatama2017generative, ding2019latent}. The fundamental idea is to treat text classification as a conditional sequence modeling task, where the classification output is determined by the label that best explains (i.e., can most accurately regenerate) the input sequence. This principle is the same as what is used for the Naive Bayes classifier, where the predicted class \( \hat{y} \) is determined by:
\begin{equation}
    \hat{y} = \arg\max_{y} p(y \mid \mathbf{x}) = \arg\max_{y} p(\mathbf{x} \mid y) p(y)
\end{equation}

\noindent Here, \( \mathbf{x} = (x_1, x_2, \ldots, x_T) \) is the observed input token sequence of length \( T \), where each token \( x_t \in \mathbb{Z}^+ \) indexes a word from a vocabulary of size \( V \). The term \( p(y \mid \mathbf{x}) \) is the posterior probability of class \( y \) given the input, which by Bayes’ rule, is proportional to the product of the class-conditional likelihood \( p(\mathbf{x} \mid y) \) and the class prior \( p(y) \). In Naive Bayes, \( p(\mathbf{x} \mid y) \) assumes feature independence and a simple parametric form. However, in the generative LSTM, this likelihood is modeled autoregressively as:
\begin{equation}\label{eqn:likelihood}
    p(\mathbf{x} \mid y) = \prod_{t=2}^{T} p(x_t \mid x_{<t}, y)
\end{equation}
where \( x_{<t} = (x_1, x_2, \ldots, x_{t-1}) \) denotes the prefix sequence up to token \( t-1 \). This formulation captures the sequential dependencies between tokens while conditioning on the class label \( y \in \{1, \ldots, C\} \), where \( C \) is the total number of output classes.

The generative model architecture consists of four main components: (i) an input embedding layer, (ii) an LSTM sequence generator, (iii) a label embedding layer, and (iv) a linear decoder. Let \( E \in \mathbb{R}^{V \times d} \) denote the token embedding matrix, where \( d \) is the dimensionality of each embedded token vector. Each input token \( x_t \) is mapped to a dense vector:
\begin{equation}
    \mathbf{e}_t = E[x_t] \in \mathbb{R}^d
\end{equation}

Similarly, class labels are embedded using a separate label embedding matrix \( L \in \mathbb{R}^{C \times d_\ell} \), where \( d_\ell \) is the dimensionality of the label embedding. For a given class \( y \), the label embedding is:
\begin{equation}
    \mathbf{l}_y = L[y] \in \mathbb{R}^{d_\ell}
\end{equation}

\noindent The embedded input sequence \( (\mathbf{e}_1, \ldots, \mathbf{e}_T) \) is passed through a stack of \( L \) LSTM layers over \( T \) time steps. For each time step \( t \in \{1, \ldots, T\} \) and each layer \( l \in \{1, \ldots, L\} \), the LSTM state is updated as:
\begin{equation}
    \mathbf{s}_t^{(l)} = F^{(l)}(\mathbf{s}_{t-1}^{(l)}, \mathbf{u}_t^{(l)})
\end{equation}

\noindent where the input to each layer $l$ is defined as:
\[
{\bf u}_t^{(l)} =
\begin{cases}
{\bf e}_t & \text{if } l = 1 \\
{\bf h}_t^{(l-1)} & \text{if } l > 1
\end{cases}
\]

\noindent and the combined state ${\bf s}_{t}^{(l)}$ includes the hidden and cell states at layer $l$ and time step $t$ as below:
$$
{\bf s}_{t}^{(l)} = ({\bf h}_{t}^{(l)}, {\bf c}_{t}^{(l)})
$$

\noindent To compute the predictive distribution for the next token \( x_{t+1} \), the hidden state \( \mathbf{h}_t \) is concatenated with the label embedding \( \mathbf{l}_y \), and passed through a fully connected linear output layer with weight matrix \( W \in \mathbb{R}^{V \times (h + d_\ell)} \):
\begin{equation}
    \mathbf{z}_t = W \cdot [\mathbf{h}_t; \mathbf{l}_y]
\end{equation}

\noindent The resulting vector \( \mathbf{z}_t \in \mathbb{R}^V \) represents the logits (unnormalized log probabilities) over the vocabulary for predicting the next token \( x_{t+1} \). \\

To compute the log-likelihood term $\log p(x_{t+1} \mid x_{\leq t}, y)$, the model first converts the unnormalized logits $\mathbf{z}_t$ into probabilities using the softmax function:
\begin{equation}
    p(x_{t+1} = i \mid x_{\leq t}, y) = \frac{\exp(\mathbf{z}_{t,i})}{\sum_{j=1}^{V} \exp(\mathbf{z}_{t,j})}
\end{equation}

\noindent
where $i \in \{1, \dots, V\}$ denotes the index of the ground truth token at position $t + 1$, and $j$ is the index over the vocabulary. The numerator selects the exponentiated logit for the correct token, while the denominator sums over all token logits to normalize. 

The generative LSTM classifier is trained using the cross-entropy loss over the next-token prediction. Given the true input sequence \( x = (x_1, \ldots, x_T) \), the objective is to maximize the log-likelihood of this sequence conditioned on class \( y \), or equivalently, to minimize the negative log-likelihood:
\begin{equation}
    \mathcal{L}_{\text{gen}} = - \sum_{t=1}^{T-1} \log p(x_{t+1} \mid x_{\leq t}, y)
\end{equation}

\noindent At inference time, the generative model evaluates \( \mathcal{L}_{\text{gen}} \) for each class \( y \in \{1, \ldots, C\} \), and selects the class that yields the lowest generation loss:
\begin{equation}
    \hat{y} = \arg\min_{y \in \{1, \ldots, C\}} \mathcal{L}_{\text{gen}}(x, y)
\end{equation}

\noindent This classification strategy leverages the model's ability to assess how well a class-conditioned sequence generator can reconstruct the given input which first involves computing \( p(x \mid y) \) and then selecting the class with the highest likelihood.

\section{Quantization}

Quantization is a model compression technique that aims to reduce the computational and memory requirements of deep neural networks by converting high-precision parameters and activations typically stored in 32-bit floating-point format into lower-precision fixed-point or integer representations. This process allows models to be executed on energy-efficient hardware such as microcontrollers, mobile processors, and embedded accelerators, where resource constraints prohibit the use of full-precision arithmetic. Quantization not only reduces the size of the model but also allows for faster inference and lower power consumption, making it a key enabler for deploying deep learning systems on edge devices. There are two principal approaches to neural network quantization: Quantization-Aware Training (QAT) and Post-Training Quantization (PTQ). These are described in the sub-sections below.

\subsection{Quantization-Aware Training (QAT)}

QAT is a technique where quantization effects are incorporated into the training loop by incorporating low-precision arithmetic during weight and bias updates.
This allows the network to learn parameters that are robust to the effects of low-precision representation \cite{zhang2022learning, esser2019learned,nagel2021}. QAT typically achieves superior accuracy compared to PTQ, especially at very low bit-widths. However, it requires optimization with full access to the training data including labels, and often requires more computational resources and higher runtimes. In scenarios where retraining is not feasible, QAT is not a viable option.

\subsection{Post-Training Quantization (PTQ)}

Post-Training Quantization (PTQ) involves applying quantization to a model after it has been fully trained, without requiring access to training gradients or re-optimization. Compared to Quantization-Aware Training (QAT), PTQ offers a simpler and more practical approach for model size and inference latency reduction, especially in scenarios where retraining is infeasible due to limitations arising from data availability, or because of compute constraints.

In this study, we adopt PTQ due to its ease of applicability in practical scenarios. It can be applied directly to a trained model and only requires a small unlabeled dataset for collecting activation statistics which are used to perform activation quantization. It is important to distinguish between how PTQ handles weights and activations: weights are quantized statically during the initial quantization step i.e. their quantized values are fixed before performing inference. In contrast, activations are not quantized in this step and instead, placeholder quantization modules are inserted for activations, and related parameters such as scales and zero points are initialized but remain unset. Using a small set of representative data, these parameters are determined later by collecting activation statistics in a step termed as {\bf calibration}. The finalized parameters are then used to quantize the activation values on the fly during inference.

\begin{figure}[H]
\centering
\begin{tikzpicture}[
    block/.style={draw, fill=blue!20, minimum width=3.8cm, minimum height=1cm, align=center, rounded corners},
    arr/.style={thick, ->, >=Stealth},
    dashedbox/.style={draw=black, dashed, inner sep=0.3cm, rounded corners}
]

\node[block] (fullmodel) {Trained Full\\Precision Model};
\node[block, below=1cm of fullmodel] (raw) {Raw Quantization};
\node[block, below=0.8cm of raw] (calib) {Calibration};
\node[block, below=0.8cm of calib] (gpfq) {GPFQ};

\node[dashedbox, fit=(raw)(calib)(gpfq), label=right:PTQ] (box) {};

\draw[arr] (fullmodel) -- (raw);
\draw[arr] (raw) -- (calib);
\draw[arr] (calib) -- (gpfq);

\end{tikzpicture}

    \caption{Post-training quantization workflow. Full Precision Model is trained, followed by quantization, calibration, and GPFQ.}
    \label{fig:quant_flow}
\end{figure}
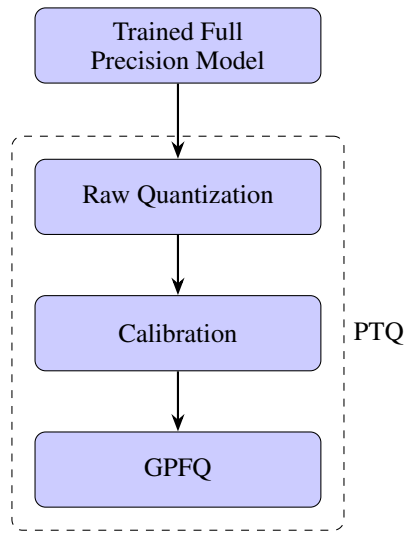

Overall the PTQ process used in our work consists of three main phases: (1) quantization of weight tensors, (2) collecting activation statistics {(calibration)} for quantizing activations during inference, (3) further improving accuracy of the quantized model using Gradient Path Following Quantization (GPFQ) method. 
The full PTQ flow in this work is shown in Figure \ref{fig:quant_flow}. These steps are described below.

\subsubsection{Weight Quantization and Activation Parameter Initialization}
\label{sec:raw quantization}

The first step in PTQ is the static quantization of the trained full-precision weight matrices. At the same time, quantization modules for activations are inserted into the model, with their parameters (e.g., scale and zero-point) left uninitialized. This sets up the infrastructure needed for runtime activation quantization, which will be finalized with the activation statistics. 
The mathematical setup for quantization of weights and activations is described briefly as follows.\\

Let \( \mathbf{W} \in \mathbb{R}^{m \times n} \) represent a full-precision weight matrix in the model. The quantized weight matrix \( \tilde{\mathbf{W}} \) is computed as:
\begin{equation}
    \tilde{\mathbf{W}} = \text{clip}\left(\left\lfloor \frac{\mathbf{W}}{\Delta_w} +z_w \right\rceil, q_{\text{min}}, q_{\text{max}} \right) \cdot \Delta_w
\end{equation}

\begin{equation}\label{eqn:zero_weight}
    z_w = \left\lfloor -\frac{\min(\mathbf{W})}{\Delta_w} \right\rceil
\end{equation}

Here, \( \Delta_w \) is the quantization step size (also known as the scale factor), and \( q_{\text{min}} \), \( q_{\text{max}} \) define the quantization range (e.g., \([-127, 127]\) for signed 8-bit integers). \( z_w \) is the zero point which is set to zero for symmetric quantization and computed using Equation~\ref{eqn:zero_weight} for asymmetric quantization. The clip function ensures that the values remain within the allowed representational bounds. The scale factor \( \Delta_w \) is typically computed using either the min-max method or a percentile-based method. The min-max method calculates the scale based on the absolute minimum and maximum values of the weight tensor, ensuring full coverage of its dynamic range as per the given equation. 
\begin{equation}
    \Delta_w = \frac{\max(\mathbf{W}) - \min(\mathbf{W})}{q_{\text{max}} - q_{\text{min}}}
\end{equation}

\noindent While simple, this approach can be sensitive to outliers. In contrast, the percentile-based method computes the range by excluding a small fraction of extreme values at both ends of the distribution (e.g., clipping to the 0.01st and 99.99th percentiles). This yields a more robust estimate of the effective value range, particularly useful for cases where rare outliers may otherwise distort the scale. In addition to weights, activation values \( \mathbf{A} \) are quantized on the fly during inference as:
\begin{equation}
    \tilde{\mathbf{A}} = \text{clip}\left(\left\lfloor \frac{\mathbf{A}}{\Delta_a}+ z_a \right\rceil, q_{\text{min}}, q_{\text{max}} \right) \cdot \Delta_a
\end{equation}

Here $\tilde{\mathbf{A}}$ denotes the quantized activation values and \( z_a\) denotes the zero point.
In this step of PTQ, quantization modules for activations are inserted into the
model, with their parameters \( \Delta_a\) and \( z_a\) left uninitialized. The actual values for \( \Delta_a\) and \( z_a\) are determined in the calibration step of PTQ.

\subsubsection{Calibration}\label{sec:calibration}

Calibration is a fundamental component of the PTQ process. Its primary purpose is to estimate the statistical distribution of intermediate activations within a neural network, enabling the derivation of appropriate activation quantization parameters. These parameters include the activation quantization scale factor \( \Delta_a\) and a zero point \( z_a\). Unlike training, calibration is a gradient-free process that typically operates on a small, fixed subset of the training data. Its accuracy is essential for minimizing quantization-induced degradation in model performance.

Mathematically, calibration involves calculating the dynamic range of a given activation tensor \( \mathbf{A} \) and mapping it to a discrete set of quantized values. 
Similar to \(\Delta_w\) the scale factor \( \Delta_a \) is typically determined using either min-max range or percentile-based statistics from activation.  The equation in the min-max case is as shown below: 

\begin{equation}\label{eqn:act_range}
    \Delta_a = \frac{\max(\mathbf{A}) - \min(\mathbf{A})}{q_{\text{max}} - q_{\text{min}}}
\end{equation}

For symmetric quantization, zero point \( z_a \) is set to zero and if asymmetric quantization is used, the zero point \( z_a \) is computed as:
\begin{equation}
    z_a = \left\lfloor -\frac{\min(\mathbf{A})}{\Delta_a} \right\rceil
\end{equation}
This allows the mapping of real-valued activations to quantized integers via:
\begin{equation}
    \tilde{a} = \text{clip}\left(\left\lfloor \frac{a}{\Delta_a} + z_a \right\rceil, q_{\text{min}}, q_{\text{max}} \right)
\end{equation}
These quantized values are then used during inference to minimize the memory and computational footprint of the network, especially on resource-constrained hardware.

\subsubsection{Greedy Path-Following Quantization (GPFQ)}\label{sec:gpfq}

While post-training quantization (PTQ) upto calibration adjusts scale factors and zero-points of weights and activations to reduce quantization errors, it does not explicitly optimize the model’s outputs to match those of the full precision model. Consequently, output mismatches may persist particularly at low bitwidths due to insufficient alignment between the quantized weights and the full-precision weight distributions.

To address this limitation, we adopt \textit{Greedy Path-Following Quantization (GPFQ)}~\cite{lybrand2021greedy} as a refinement step following calibration. GPFQ is a data-dependent, post-training quantization algorithm that improves output fidelity by minimizing quantization-induced forward-pass error on a batch of calibration inputs. Unlike memoryless scalar quantization (MSQ), \cite{eamaz2023matrix} which quantizes each weight independently, GPFQ quantizes each \textit{row of the weight matrix} (i.e., the weight vector associated with an output unit such as a fully connected output) sequentially and in a data-aware fashion. This approach is especially effective under very low bit-width constraints (e.g., 3- or 4-bit quantization), where maintaining the integrity of activation responses is critical for preserving accuracy.

\paragraph{Problem Formulation.}

Let:
\begin{itemize}
    \item \( m \in \mathbb{N} \) be the number of calibration samples used for quantization,
    \item \( N \in \mathbb{N} \) be the input feature dimension (i.e., the number of elements in the primary input to the network),
    \item \( X \in \mathbb{R}^{m \times N} \) denote the calibration input matrix, where each row \( x_i \in \mathbb{R}^N \) represents an input sample and each column \( X_t \in \mathbb{R}^m \) corresponds to the values of the \( t \)-th feature across all samples,
    \item \( w \in \mathbb{R}^{N_l} \) be a full-precision weight vector feeding into a given unit at layer \(l\) representing the strengths of connecting from preceeding unit to that unit. Typically we have \( {N_l}\leq N \), 
    \item \( q \in \mathcal{A}^{N_l} \) be the quantized version of \( w \), where \( \mathcal{A} \subset \mathbb{R} \) is a finite quantization alphabet (e.g., \( \{-1, 0, 1\} \)).
\end{itemize}

\noindent The objective of GPFQ is to find $q$ that minimizes the squared output error:
\begin{equation}
\min_{q \in \mathcal{A}^{N_l}} \|Xw - \widetilde{X}q\|_2^2
\end{equation}
where $\widetilde{X}$ is the (potentially quantized) activations from the previously quantized layers.
If all weights are considered simultaneously across all layers then this is a combinatorial optimization problem and is known to be \textit{NP-hard} due to the exponential number of possible quantized vectors $q$ across all layers. Instead of solving it exactly, GPFQ uses a greedy sequential strategy that approximates a near-optimal solution efficiently.


\subsection{Methodology}


\subsubsection{Data Preprocessing and Model Training}

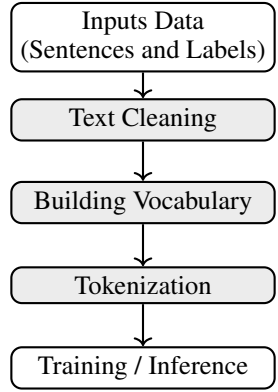
\begin{figure}[h]
    \centering
    \centering
    \begin{tikzpicture}[node distance=1.1cm, every node/.style={align=center}]
    
    \node (input) [draw, rounded corners, thick, minimum width=3.2cm, minimum height=0.9cm, minimum width=3.5cm] {Inputs Data\\(Sentences and Labels)};
    \node (clean) [draw, rounded corners, thick, fill=gray!15, below of=input, minimum width=3.5cm] {Text Cleaning};
    \node (vocab) [draw, rounded corners, thick, fill=gray!15, below of=clean, minimum width=3.5cm] {Building Vocabulary};
    \node (token) [draw, rounded corners, thick, fill=gray!15, below of=vocab, minimum width=3.5cm] {Tokenization};
    \node (train) [draw, rounded corners, thick, below of=token, minimum width=3.5cm] {Training / Inference};
    
    \draw[->, thick] (input) -- (clean);
    \draw[->, thick] (clean) -- (vocab);
    \draw[->, thick] (vocab) -- (token);
    \draw[->, thick] (token) -- (train);
    
    \end{tikzpicture}
    \caption{Text preprocessing pipeline for both generative and discriminative classifiers.}
    \label{fig:preprocessing_diagram}
\end{figure}

We establish baselines for our quantization experiments by training full precision discriminative and generative LSTM models based on the architectures described in \cite{yogatama2017generative}. Prior to model training, all text data are preprocessed as  shown in Figure~\ref{fig:preprocessing_diagram} where the input sentences are first cleaned to remove punctuation symbols, special characters, and excess whitespaces. The cleaned text is then used to construct a vocabulary based on word frequency, and each word or token of a given sentence is mapped to a sequence of integers using Spacy Tokenization\cite{honnibal2020spacy}. These integer representations form the input to the LSTM models both during  training and inference.

\begin{figure}[h]
\centering
\begin{tikzpicture}[node distance=1cm, every node/.style={align=center}]
    \node (input) [draw, rounded corners, thick, minimum width=3.5cm]{Text Inputs\\(Pre-processed)};
    \node (embed) [draw, rounded corners, thick, below of=input, fill=blue!20, minimum width=3.5cm] {Embedding Layer};
    \node  (lstm) [draw, rounded corners, thick, below of=embed, fill=blue!20, minimum width=3.5cm] {LSTM Layer};
    \node  (fc) [draw, rounded corners, thick, below of=lstm, fill=blue!20, minimum width=3.5cm] {Fully Connected Layer};
    \node (logit) [draw, rounded corners, thick, below of=fc, minimum width=3.5cm] {Logit Outputs for Labels};
    \node (argmax) [draw, rounded corners, thick, below of=logit, minimum width=3.5cm] {Argmax (over $C$ classes)};
    \node (output) [draw, rounded corners, thick, below of=argmax, minimum width=3.5cm] {Predicted Label};

    \draw[->, thick] (input) -- (embed);
    \draw[->, thick] (embed) -- (lstm);
    \draw[->, thick] (lstm) -- (fc);
    \draw[->, thick] (fc) -- (logit);
    \draw[->, thick] (logit) -- (argmax);
    \draw[->, thick] (argmax) -- (output);
\end{tikzpicture}
\caption{Dataflow in the discriminative LSTM classifier during inference. Trainable layers are shaded in blue.}
\label{fig:disc_model}
\end{figure}
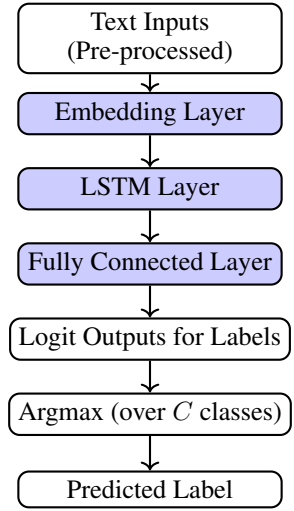

The dataflow of the discriminative LSTM classifier during inference is depicted in Figure~\ref{fig:disc_model}. The input sequence is first passed through an embedding layer to convert tokens into dense vector representations. These embeddings are then processed by the LSTM layer and following this the hidden state of the final layer of the LSTM is fed into a fully connected layer, producing a logit vector corresponding to class scores. A softmax layer transforms these logits into class probabilities, and the predicted label is the one that minimizes the cross-entropy loss with respect to these estimated probabilities.

\begin{figure}[h]
\centering
\begin{tikzpicture}[node distance=1cm, every node/.style={align=center}]
    \node (text) [draw, rounded corners, thick, minimum width=3.3cm] at (-4, 0) {Text Inputs\\(Pre-processed)};
    \node (label) [draw, rounded corners, thick, minimum width=3.3cm] at (0, 0) {Class Labels\\(1, 2, ..., C)};
    
    \node (wordemb) [draw, rounded corners, thick, below of=text, fill=blue!20] {Word Embedding Layer};
    \node (labelemb) [draw, rounded corners, thick, below of=label, fill=blue!20] {Label Embedding Layer};
    
    \node (lstm) [draw, rounded corners, thick, below of=wordemb, fill=blue!20, minimum width=3.5cm] {LSTM Layer};
    \node (fc) [draw, rounded corners, thick, below of=lstm, fill=blue!20, minimum width=3.5cm] {Fully Connected Layer};
    \node (logits) [draw, rounded corners, thick, below of=fc, minimum width=3.5cm] {Class Conditioned Logit Outputs\\for Predicted Words};
    \node (loss) [draw, rounded corners, thick, below=.3cm of logits, minimum width=3.5cm] {Cross Entropy Losses \\for Predicted Words};
    \node (argmin) [draw, rounded corners, thick, below of=loss, minimum width=3.5cm] {Argmin (over $C$ classes)};
    \node (output) [draw, rounded corners, thick, below of=argmin, minimum width=3.5cm] {Predicted Label};

    \draw[->, thick] (text) -- (wordemb);
    \draw[->, thick] (label) -- (labelemb);
    \draw[->, thick] (wordemb) -- (lstm);
    \draw[->, thick] (labelemb) |- (fc);
    \draw[->, thick] (lstm) -- (fc);
    \draw[->, thick] (fc) -- (logits);
    \draw[->, thick] (logits) -- (loss);
    \draw[->, thick] (loss) -- (argmin);
    \draw[->, thick] (argmin) -- (output);
    \draw[->, thick] (text.west) -- ++(-1.0, 0) 
              -- ++(0, -5.2)       
              -- (loss.west);  
\end{tikzpicture}
\caption{Dataflow in the generative LSTM classifier during inference. Trainable layers are shaded in blue.}
\label{fig:gen_model}
\end{figure}
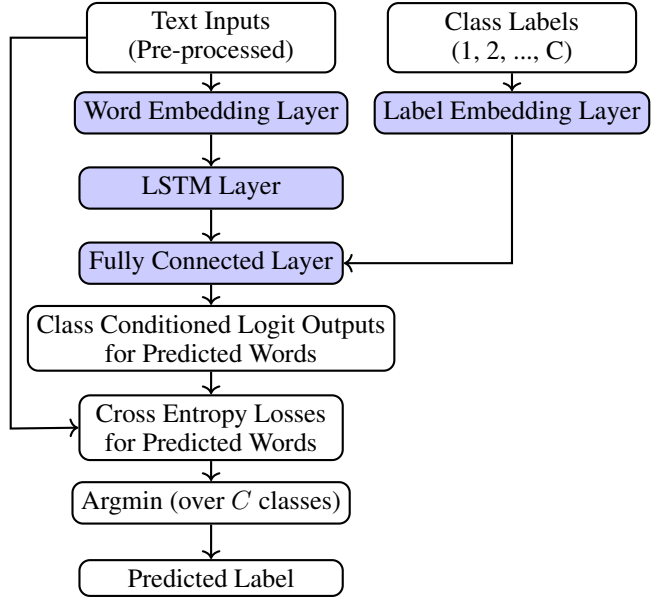

Figure~\ref{fig:gen_model} shows the dataflow of the generative LSTM classifier during inference. 
In this case, the input sequence is processed through the word embedding layer in the same manner as the discriminative model. Following this, the LSTM forecasts the $t$-th word in the input sequence given a class label $c \in \{1, \ldots, C\}$ and previous $1, \ldots, (t-1)$ words where $t \in \{1, \ldots, T\}$. Here, $C$ denotes the total number of class labels and $T$ is the length of the input sequence. Using these predicted words and ground truth words from the input sequence, cross-entropy loss values are calculated for the $C$ classes. Finally, the predicted value is determined by the label that has the minimum cross entropy among all classes. In this manner, the generative LSTM performs classification by autoregressively reconstructing the input sequence conditioned on the class labels.  

In this work both models are trained using a single-layer LSTM configuration with 100 hidden units and an embedding dimension of 100 following the model configurations given in \cite{yogatama2017generative, ding2019latent} We used a batch size of 32 and an Adam optimizer with a learning rate of 0.001. Training is continued until convergence is observed on a held-out validation set. 
\subsubsection{Quantization Process Flow}

The post-training quantization of both generative and discriminative classifiers is performed using the Brevitas framework \cite{brevitas}. 
The PTQ flow in Brevitas begins with defining the model architecture and loading pre-trained full-precision parameters. This is followed by preprocessing the input text data to match the format used during training following which quantization is applied on all trainable layers across the entire model, including the embedding, LSTM and fully connected layers.

Quantization is carried out using symmetric quantization for both weights and activations. All tensor values within a layer are quantized using a single shared scale factor and zero point thereby maintaining per-tensor granularity. The bit-widths for both weights and activations are varied across experiments from 8 bits down to 3 bits to analyze the impact of increasingly aggressive quantization. Bias values which do not significant impact on the size of the model are excluded from quantization and retained in full 32-bit floating-point precision. 

The scale factors for quantization are computed using statistical analysis of weight and activation values observed during calibration following the process described in Section \ref{sec:calibration}. For activation quantization, a percentile-based method is employed, clipping extreme values beyond the 99.99th percentile to reduce the influence of outliers. All quantization operations are performed in integer format, and floating-point scale factors are used to preserve precision in the quantization-dequantization mapping. 

Following initial quantization, we perform calibration using a small subset of the training data. This calibration set is used to estimate activation distributions and finalize the scale parameters for each layer. After calibration, we apply Greedy Path-Following Quantization (GPFQ), which iteratively adjusts weight values to minimize quantization error and improve inference accuracy, particularly in low bitwidth configurations as described in Section~\ref{sec:gpfq}. In our implementation, GPFQ is applied only to the final linear (fully connected) layer. Due to the internal recurrence and gating operations in LSTM modules, Brevitas does not currently support GPFQ for LSTM layers. Consequently, weights in the LSTM remain quantized but unrefined by GPFQ for all our experiments. It should be noted that no labels are required during both calibration and GPFQ phases of the PTQ process. This comprehensive quantization strategy combining symmetric integer quantization, percentile-based calibration, and GPFQ enables precise control over the trade-off between model compression and performance. The use of consistent parameter settings across both model types ensured a fair comparison of quantization robustness. 

\begin{figure}[h]
    \centering
    \includegraphics[width=0.35\linewidth]{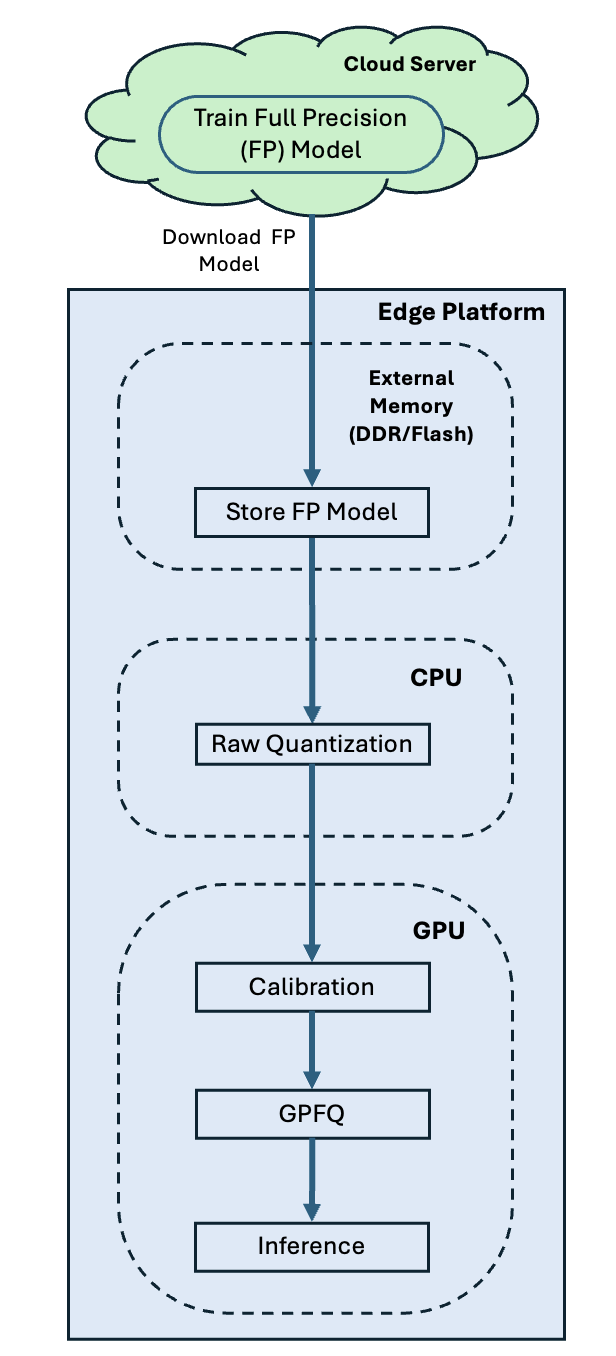}
    \caption{Deployment workflow for post-training quantization. The FP model is trained offline/in cloud, after which it is downloaded to an edge device. Quantization, calibration, and GPFQ are applied on-device prior to inference.}
    \label{fig:quant_architecht}
\end{figure}

Figure~\ref{fig:quant_architecht} illustrates the deployment workflow for the PTQ process discussed in this work. The full-precision (FP) model is first trained on a high-resource cloud server and then downloaded to the edge device for deployment. On the edge platform, the FP model is initially stored in external memory (e.g., DDR or Flash). The PTQ process begins with raw quantization, followed by calibration using unlabeled data, and then refinement using GPFQ. The calibration and GPFQ steps are executed on the device GPU to adapt the model for low precision inference. Once quantization is complete, the optimized model is used for real-time inference on the GPU. This architecture reflects realistic deployment constraints, where on-device resources are limited and retraining is impractical, requiring all quantization steps to be performed post hoc.

\subsubsection{Noise Robustness}\label{subsec:noise_robustness_method}

To evaluate the resilience of the models under noisy input conditions, we conducted a series of noise injection experiments on both the full-precision and quantized versions of the generative and discriminative classifiers. These experiments are designed to simulate real-world input corruption such as typographical errors, OCR artifacts, or transmission noise common in edge-deployed natural language systems.

The noise is introduced by applying random character-level replacements to the test dataset. Specifically, for each character in the input sentence, a substitution is performed with a probability \( \epsilon \in [0, 0.1] \), where the replacement character is sampled uniformly from the model's vocabulary. This controlled corruption mechanism allows us to systematically assess how inference accuracy deteriorates with increasing levels of noise as reflected by increasing \( \epsilon\).

The experiment is carried out in two stages. First, the original full-precision generative and discriminative classifiers are evaluated under varying noise levels to establish baseline robustness. Accuracy is measured at each noise level and compared across models. In the second stage, the same noise injection process is applied to inputs fed into quantized models across a range of bit-widths (3-bit to 8-bit), This enables a direct comparison of how quantization interacts with noise, particularly in low bitwidth scenarios where representational capacity is limited.


\section{Experiments and Results}


\subsection{Dataset Description}

To evaluate the performance of our generative and discriminative text classification models, we conduct experiments on two benchmark datasets: \textbf{AG News} and \textbf{DBPedia}.

\begin{itemize}
    \item \textbf{AG News} is a widely used dataset for topic classification, consisting of news articles categorized into four classes: World, Sports, Business, and Science/Technology. The dataset contains 120,000 training samples and 7,600 test samples \cite{zhang2015}.
    
    \item \textbf{DBPedia} is a large-scale ontology classification dataset derived from Wikipedia, consisting of 14 non-overlapping classes. It includes 560,000 training samples and 70,000 test samples, with each sample containing a title and abstract extracted from a DBPedia article.
\end{itemize}

\noindent We have taken 80\% of the trianing dataset for training the full precision model training, and 20\% of the training dataset as validation dataset.

\subsection{Full-Precision Model Evaluation}

In the first experiment, we evaluate the baseline performance of both generative and discriminative classifiers in their full-precision form, that is, without any applied quantization. Each model is trained on the respective training set and evaluated on the full test set.

The classification accuracy on the test data for each model is shown in Table~\ref{tab:fp_results}. It can ne seen that the discriminative classifier slightly outperforms the generative classifier on both datasets. However, the generative classifier achieves competitive performance and demonstrates its potential as a viable alternative.

\begin{table}[h]
\centering
\caption{Test Accuracy (\%) of Full-Precision Generative and discriminative classifiers}
\label{tab:fp_results}
\begin{tabular}{|l|c|c|}
\hline
\textbf{Model} & \textbf{AG News} & \textbf{DBPedia} \\
\hline
Generative LSTM & 88.95 & 95.93 \\
Discriminative LSTM & 89.51 & 97.62 \\
\hline
\end{tabular}
\end{table}

These results establish a high-performing reference point for comparing quantized models in later sections.

\subsection{Post-Training Quantization with Class-Unconditional Calibration}\label{subsec:PTQ_UC_Result}

To evaluate the effectiveness of PTQ under practical conditions, we first apply a class-unconditional calibration strategy to both generative and discriminative classifiers. In this setting, the calibration data is sampled randomly from the training set without ensuring balanced class representation. This design simulates a practical real-world scenario where labeled data is scarce or unavailable during model deployment. The aim is to assess how well each model preserves its predictive performance across varying bit-widths when the calibration data lacks class balance.

\begin{figure}[h]
    \centering
    \includegraphics[width=0.4\linewidth]{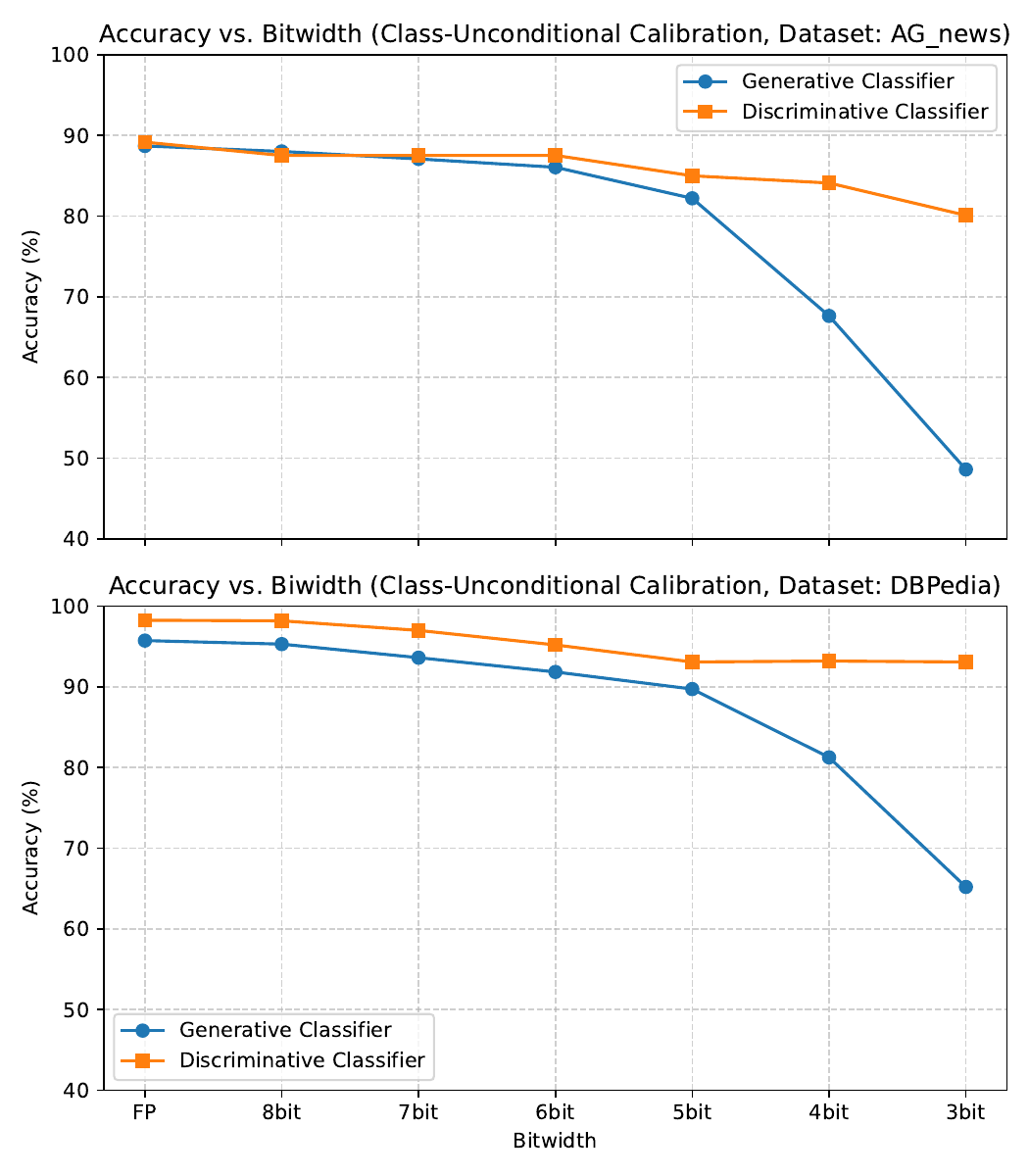}
    \caption{Quantization Accuracy of Generative and discriminative classifiers using Class-Unconditional Calibration on AG News and DBPedia datasets.}
    \label{fig:quant_acc}
\end{figure}

Quantization is implemented using the Brevitas framework, applying uniform quantization to the following layers of the network: the embedding, LSTM, and fully connected layers. We evaluate model accuracy under 8-bit, 7-bit, 6-bit, 5-bit, 4-bit, and 3-bit quantization levels. Calibration and GPFQ are performed using 25\% of the training data \cite{li2023impact}, with no class-based filtering or balancing. Increasing the calibration set size beyond 25\% did not yield noticeable improvements in performance, which motivated us to maintain the 25\% calibration proportion throughout our experiments. 

The classification results for both models across AG News and DBPedia datasets are presented in Figure~\ref{fig:quant_acc}. It can be seen that the discriminative classifier demonstrates robust behavior under quantization as its accuracy remains largely stable across higher bit-widths and only begins to show moderate degradation at 3-bit, particularly on AG News. In contrast, the generative classifier shows substantial sensitivity to quantization as its performance begins to decline noticeably below 6-bit and deteriorates rapidly at 5-bit and lower. On both datasets, the generative classifier's accuracy at 3-bit is significantly lower than its full-precision counterpart, illustrating its vulnerability to quantization when not supported by class-aware calibration.


\subsection{Post-Training Quantization with Class-Conditional Calibration}

We also examine the effectiveness of class-conditional calibration where the calibration dataset is constructed to maintain equal representation from each class in the training data. This approach ensures that the activation statistics gathered during calibration reflect the full diversity of the target label space, which is particularly important for models that rely on class-specific representations, such as generative classifiers.

The quantization process mirrors that used in earlier experiments where PTQ is applied to the embedding, LSTM, and fully connected layers using the Brevitas framework. Model performance is evaluated across a range of bit-widths from 8-bit down to 3-bit. The critical difference in this setup lies in the composition of the calibration data. Instead of being sampled randomly (as in the class-unconditional case), the calibration set is stratified so that each class contributes an equal proportion of examples e.g., 25\% from each class in a four-class classification task.

\begin{figure}[h]
    \centering
    \includegraphics[width=0.45\linewidth]{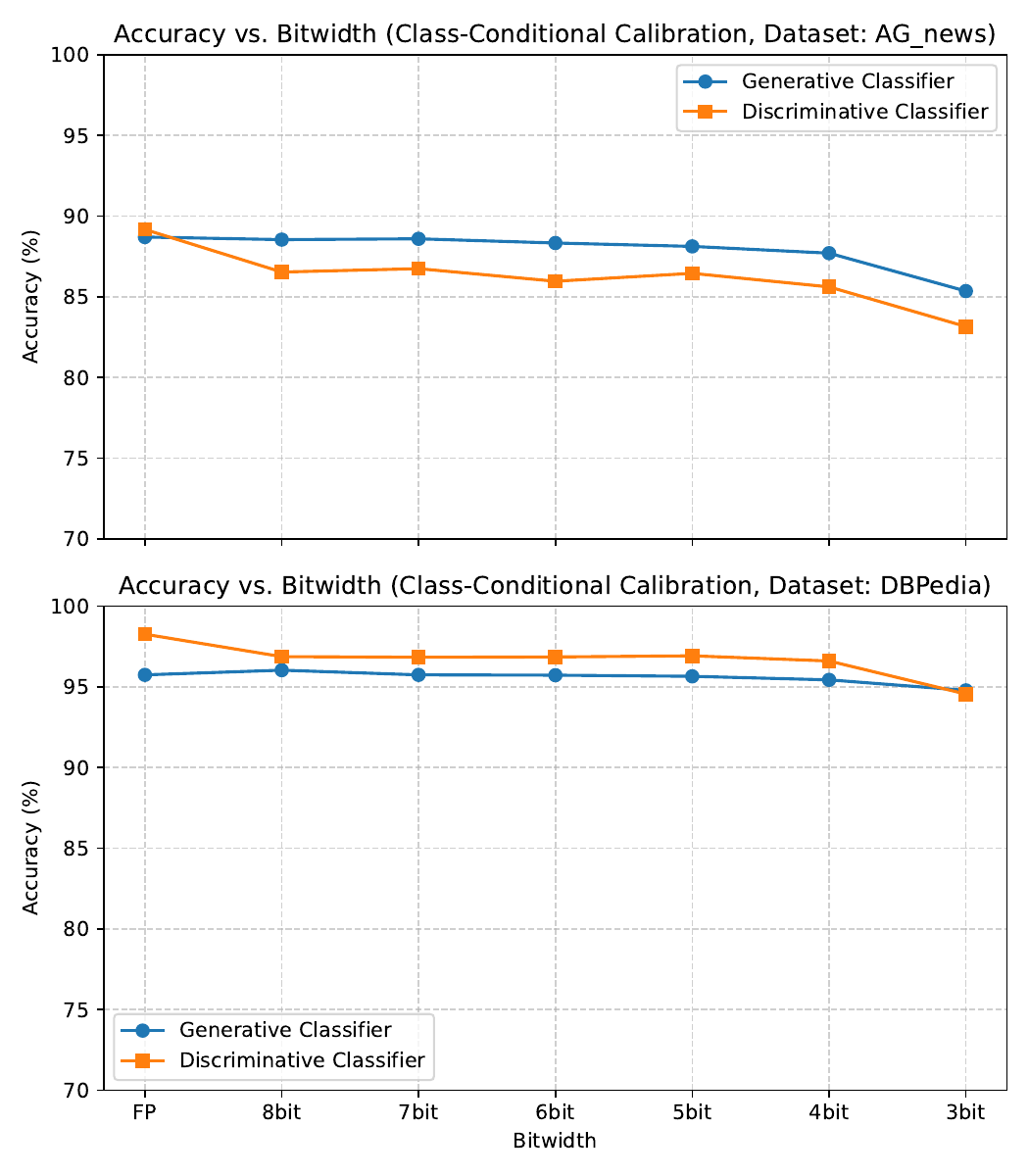}
    \caption{Quantization accuracy of generative and discriminative classifiers using class-conditional calibration on AG News and DBPedia datasets.}
    \label{fig:cc_quant_acc}
\end{figure}

\begin{figure}[h]
    \centering
    \includegraphics[width=0.45\linewidth]{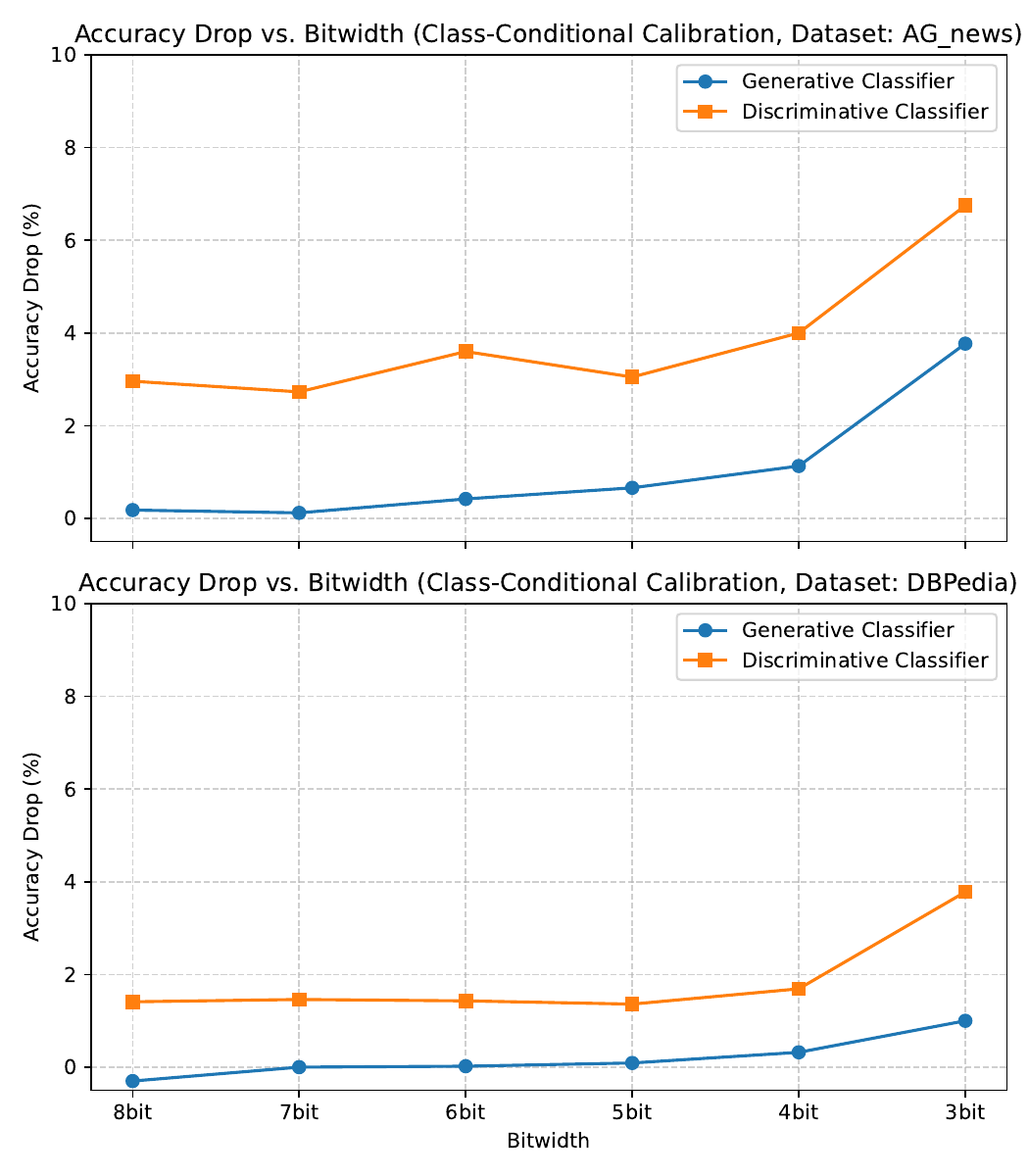}
    \caption{Accuracy drop with respect to full-precision (32-bit) model for both generative and discriminative classifiers across bit-widths, using class-conditional calibration.}
    \label{fig:cc_drop}
\end{figure}

The results of this experiment are shown in Figure~\ref{fig:cc_quant_acc}, which reports the test accuracy across bit-widths, and Figure~\ref{fig:cc_drop}, which presents the relative accuracy drop compared to the full-precision (32-bit) baseline. The generative classifier shows markedly improved resilience under class-conditional calibration. Its performance remains stable down to 4-bit and degrades only moderately at 3-bit, unlike the sharp decline observed under class-unconditional calibration. This suggests that class-aware calibration enables more effective parameter adaptation during GPFQ, helping the model to preserve consistent accuracy despite aggressive quantization.

The discriminative classifier continues to perform reliably across bit-widths, although class-conditional calibration yields slightly more stable accuracy at lower precisions. Interestingly, in this configuration, the generative classifier outperforms the discriminative classifier at several bit-widths on both AG News and DBPedia. This reversal, relative to the class-unconditional scenario, further emphasizes the importance of class coverage in calibration for generative architectures. The maximum observed accuracy drop for the generative classifier at 3-bit is under 4\%, compared to over 7–8\% in the previous experiment, reinforcing the conclusion that calibration diversity directly impacts model robustness under quantization.

These results highlight a critical difference in resilience between model types. Generative classifiers rely more heavily on structured internal representations and sequence-level dependencies, which are more susceptible to distortions introduced by low-bit quantization. Without balanced calibration data to guide the GPFQ process, these models struggle to retain reliable performance. The findings underscore the importance of developing improved sampling strategies for calibration to mitigate quantization-induced degradation, especially in generative settings.



\subsection{Input Noise Robustness Test}\label{subsec:noise_test_Result}

This experiment builds upon the noise injection procedure described in Section~\ref{subsec:noise_robustness_method}. Using the same character-level corruption strategy, we evaluated the full-precision generative and discriminative models on the AG News and DBPedia datasets under increasing noise levels.

\begin{figure}[h]
    \centering
    \includegraphics[width=0.45\linewidth]{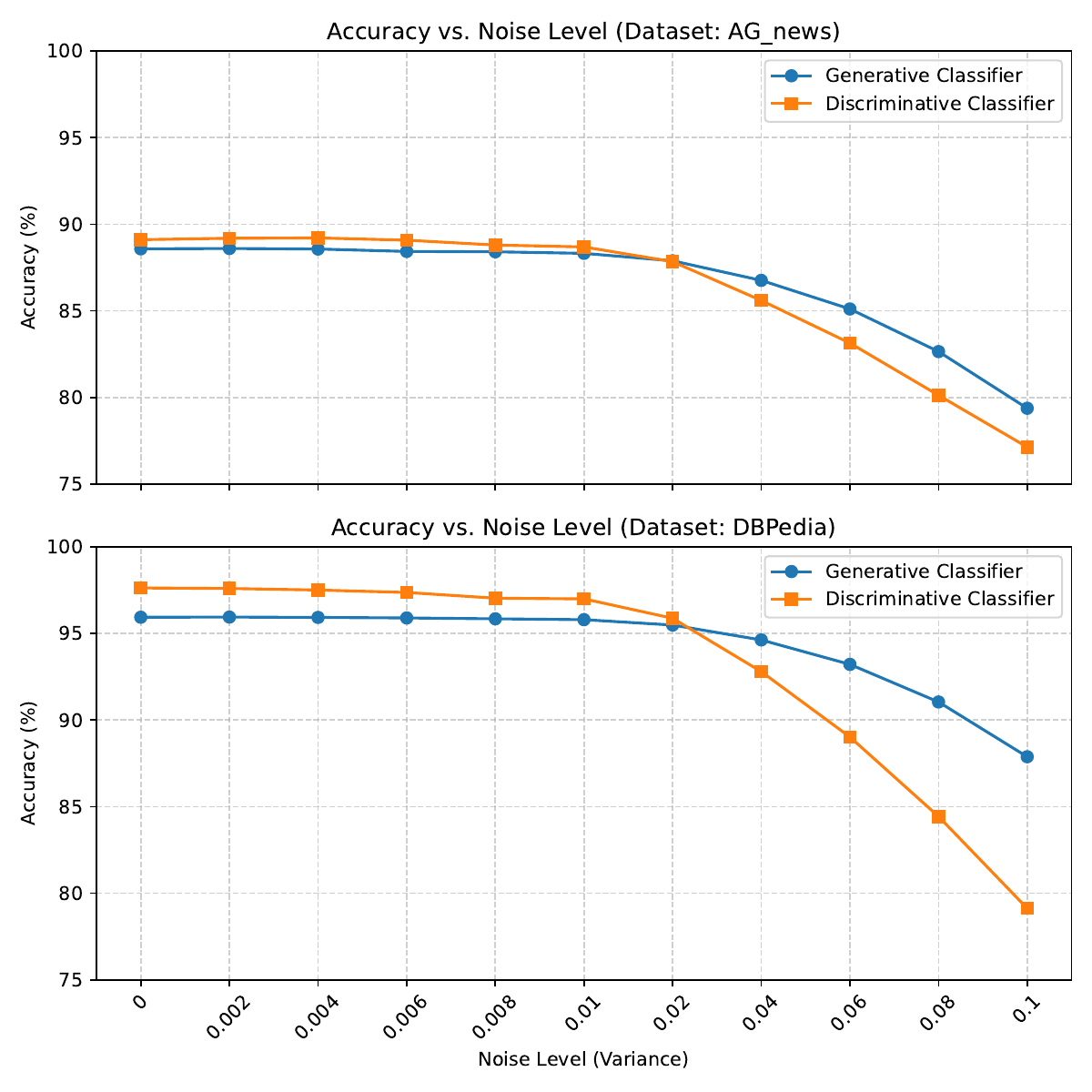}
    \caption{Accuracy vs. Noise Level for Discriminative and generative classifiers on AG News and DBPedia datasets. Noise is introduced via uniform random character substitution.}
    \label{fig:noise_robustness}
\end{figure}

For each noise level $\epsilon \in [0, 0.1]$, we measured the classification accuracy to assess the impact of input perturbations. The results quantify the resilience of both models when faced with character-level corruption, highlighting their respective robustness under realistic noisy conditions.

The results, presented in Figure~\ref{fig:noise_robustness}, show that both models maintain high performance under low noise levels. However, a clear divergence in robustness emerges as $\epsilon$ increases. The generative classifier exhibits a slower decline in accuracy compared to its discriminative counterpart, particularly beyond a noise threshold of $\epsilon = 0.02$. These findings indicate that full-precision generative LSTM classifiers offer greater robustness under text corruption as compared to full precision discriminative classifiers, making them potentially more reliable for deployment in error-prone or noisy real-world environments. These results are consistent with the findings in \cite{lee2019robust, li2018generative_robustness}, which demonstrate that generative classifiers exhibit greater robustness noise and adversarial perturbations, respectively, when compared to discriminative classifiers.


To assess how quantization influences model robustness to noisy inputs, we applied the same character-level corruption methodology described in Section~\ref{subsec:noise_robustness_method} to quantized versions of both classifiers. Each model, quantized to bit-widths ranging from 3-bit to 8-bit and calibrated using class-conditional data, was evaluated under increasing noise levels ($\epsilon \in [0, 0.1]$). This experiment provides insight into the interaction between quantization precision and input noise, highlighting the practical trade-offs in deploying low-bit models under real-world conditions where textual inputs may be imperfect.

\begin{figure}[h]
    \centering    \includegraphics[width=0.8\linewidth]{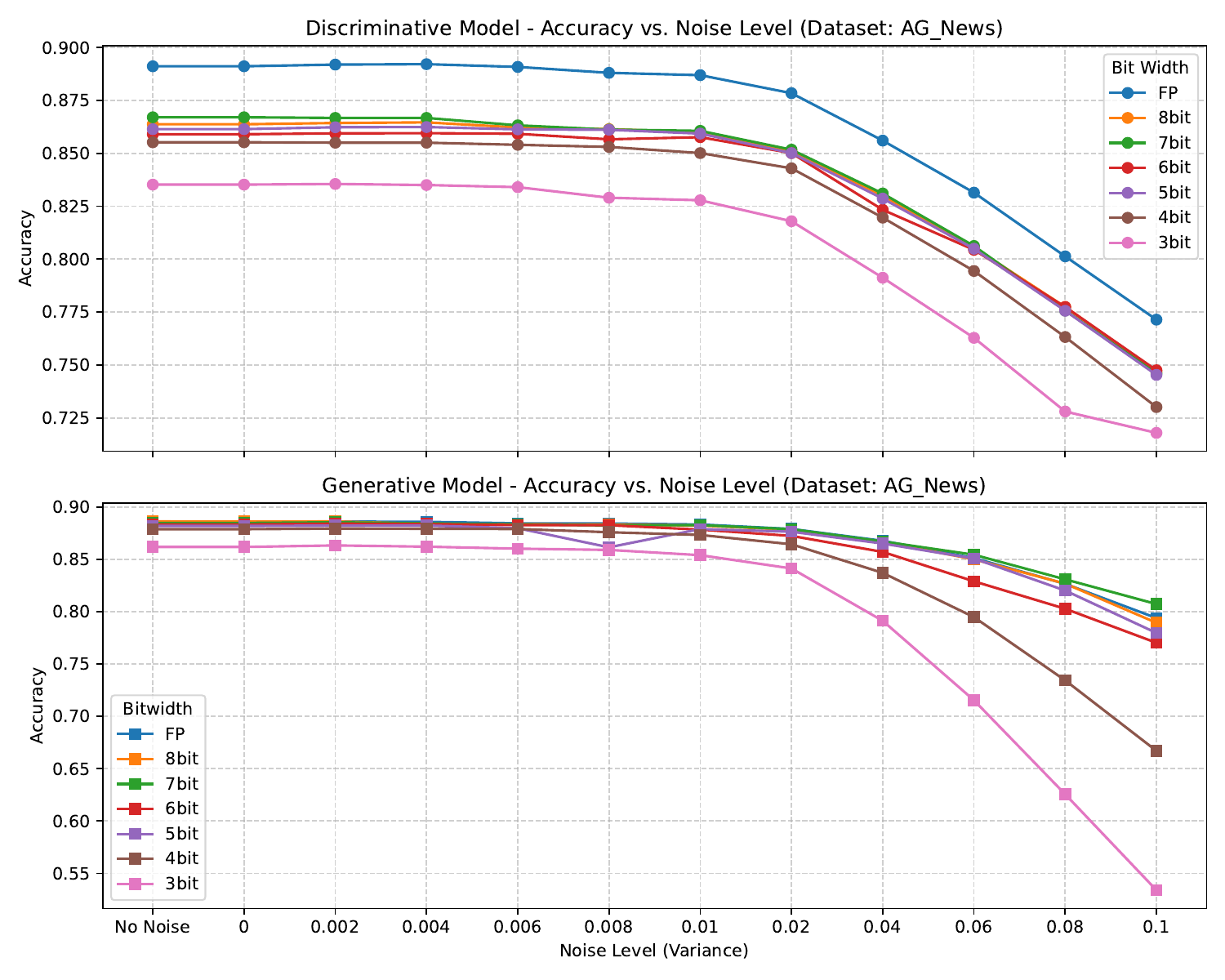}
    \caption{Noise robustness of quantized discriminative and generative classifiers. Accuracy is measured across varying input noise levels for bit-widths ranging from 3-bit to 8-bit.}
    \label{fig:quant_noise}
\end{figure}

The results are presented in Figure~\ref{fig:quant_noise}. It can be seen that the discriminative classifier displays notable resilience across all bit-widths. Its performance remains largely stable up to moderate noise levels, with only a gradual decline observed when $\epsilon$ exceeds 0.06. Even at the most aggressive quantization level (3-bit), the model retains acceptable performance, indicating that its decision boundaries are robust to both quantization noise and character-level input perturbations.

In contrast, the generative classifier exhibits significantly higher sensitivity to noise after quantization. Although the model stays on par with the discriminative baseline at higher bit-widths, accuracy begins to deteriorate sharply for $\epsilon \geq 0.04$ especially in the 4-bit and 3-bit settings. This pattern suggests that generative classifiers, which depend on sequential generation to estimate the class conditional likelihood as described by Equation~\ref{eqn:likelihood} are more vulnerable to cascading errors when the input distribution is distorted and the representation capacity is limited by quantization.

These observations point to an important trade-off between compression and robustness. While aggressive quantization can reduce memory and compute costs, it may also amplify a model's vulnerability to input corruption, particularly in generative architectures. The findings reinforce the importance of choosing bit-widths and calibration strategies carefully, especially when deploying quantized generative classifiers in real-life noise-prone environments.

We have performed the same noise-robustness experiment on models calibrated with \emph{class-unconditional} data and observed a consistent trend of performance degradation with increasing noise levels. At higher bit-widths, generative classifiers retain better robustness to noise compared to discriminative classifiers. However, as the bit-width decreases, this advantage diminishes, and discriminative models begin to outperform generative ones under noisy input conditions.


\section{Analysis and Discussion}

\subsection{Effect of Class Coverage in Calibration and GPFQ}

To further explore the role of calibration data composition in post-training quantization performance, we conducted a controlled experiment to assess how class diversity within the calibration set affects model behavior under Calibration + GPFQ. The primary motivation is to explain the significant degradation observed in generative classifiers when the calibration data lacks full class representation.

We simulated different levels of class coverage by constructing four synthetic calibration subsets derived from a 4-class classification task. The first set contains an equal number of samples from all four classes, representing a fully balanced calibration condition. The remaining subsets contain samples from only three, two, or one class(es), respectively. These sampling schemes enable a structured examination of how increasing class imbalance influences quantization quality. Each subset is used to calibrate both generative and discriminative classifiers, followed by GPFQ, and the models are evaluated across a range of bit-widths from 3-bit to 8-bit.

\begin{figure}[h]
    \centering
    \includegraphics[width=.45\linewidth]{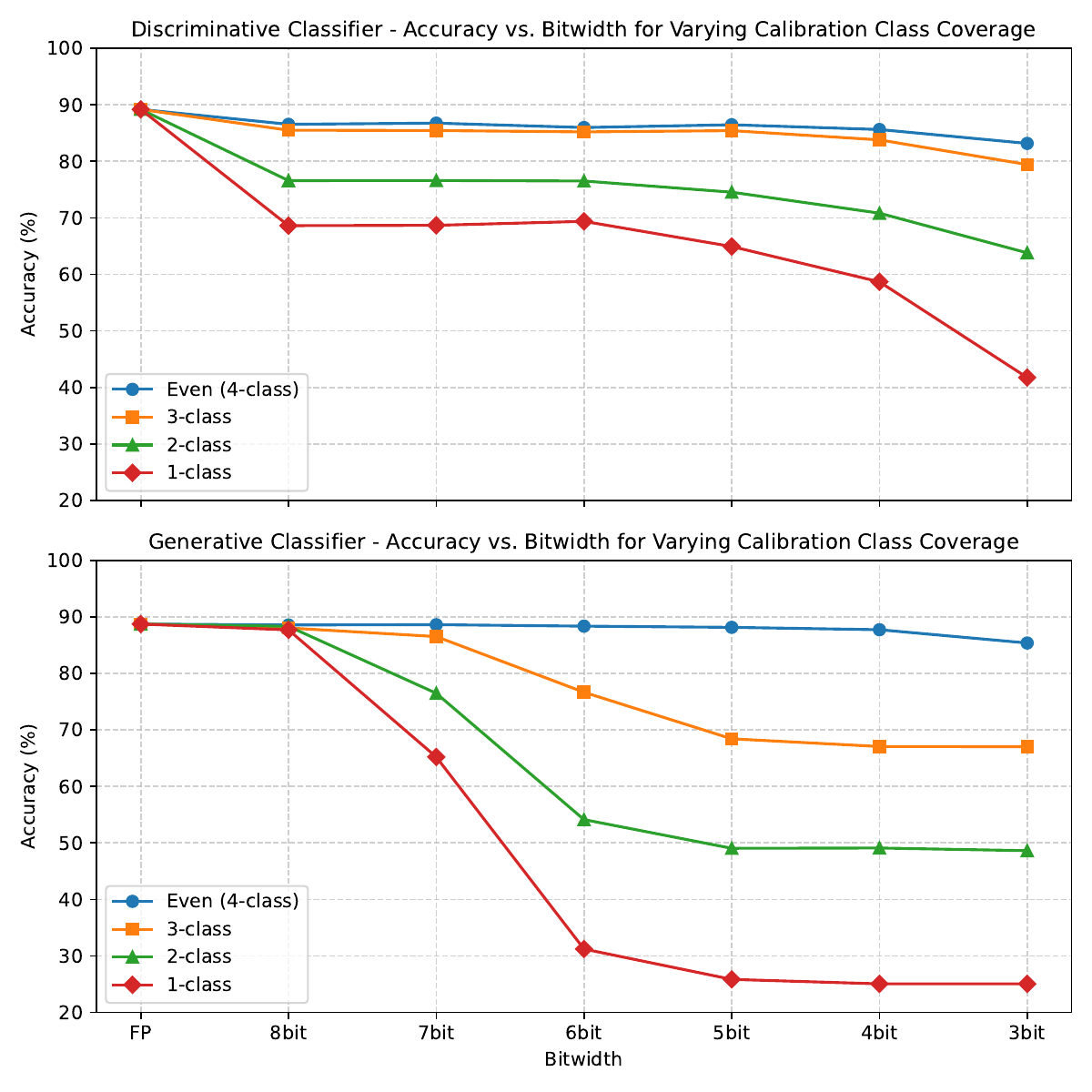}
    \caption{Accuracy of generative and discriminative classifiers under different calibration sampling schemes across bit-widths. generative classifiers show steep degradation when class diversity is lacking.}
    \label{fig:sampling_scheme}
\end{figure}

The results are shown in Figure~\ref{fig:sampling_scheme}. The generative classifier is found to be highly sensitive to class imbalance in the calibration set. When fewer classes are present particularly in the one-class and two-class schemes the model fails to collect sufficiently representative activation statistics during calibration. As a result, calibraiton and GPFQ operates with an incomplete view of the data distribution, calibration step first mis-estimates activation ranges, and the subsequent GPFQ refinement applied row-wise to trainable weight matrices of final linear layer inherits these skewed statistics, yielding scale/zero-point adjustments that are misaligned with the full data statistics (see Section~\ref{subsubsec:impact_of_cal_on_linear} and Section~\ref{subsubsec:impact_of_cal_on_act} for detailed evidence). This effect becomes increasingly detrimental as the bit-width decreases, with sharp accuracy drops observed at 5-bit and below. The degradation is most severe in the one-class calibration setting, where the generative classifier essentially loses the ability to demonstrate good performance across all classes after quantization.

In contrast, the discriminative classifier shows greater resilience under the same conditions. While its performance also declines as class diversity is reduced, the degradation is more gradual, and acceptable accuracy is retained even at 3-bit precision. This relative stability may be attributed to the model's more localized decision boundary structure, which is less reliant on label-conditioned generation than the generative LSTM.

These findings reinforce the conclusion that maintaining class coverage in the calibration data is essential for effective quantization of generative classifiers. In the absence of full class representation, the calibration process fails to prepare the model for downstream inference across all categories, leading to a breakdown in accuracy under quantization. This effect is magnified at lower bit-widths, where the room for approximation error is limited.

\subsubsection{Impact of Calibration Data on Linear Layer Weights}\label{subsubsec:impact_of_cal_on_linear}

In Section~\ref{subsec:PTQ_UC_Result} we can see that generative classifiers suffer from accuracy drop when calibtrated with class-unconditional data. To investigate why calibration with class-unconditional data causes a marked accuracy drop, we perform a comparative analysis of two variants of the quantized generative classifier: one calibrated with class-conditional data and another with class-unconditional data. Both models undergo PTQ followed by GPFQ. We then assess the effect of calibration diversity by analyzing how GPFQ shifts the weight distributions relative to the pre-calibrated full-precision model.

\begin{figure}[h]
    \centering
    \includegraphics[width=0.45\linewidth]{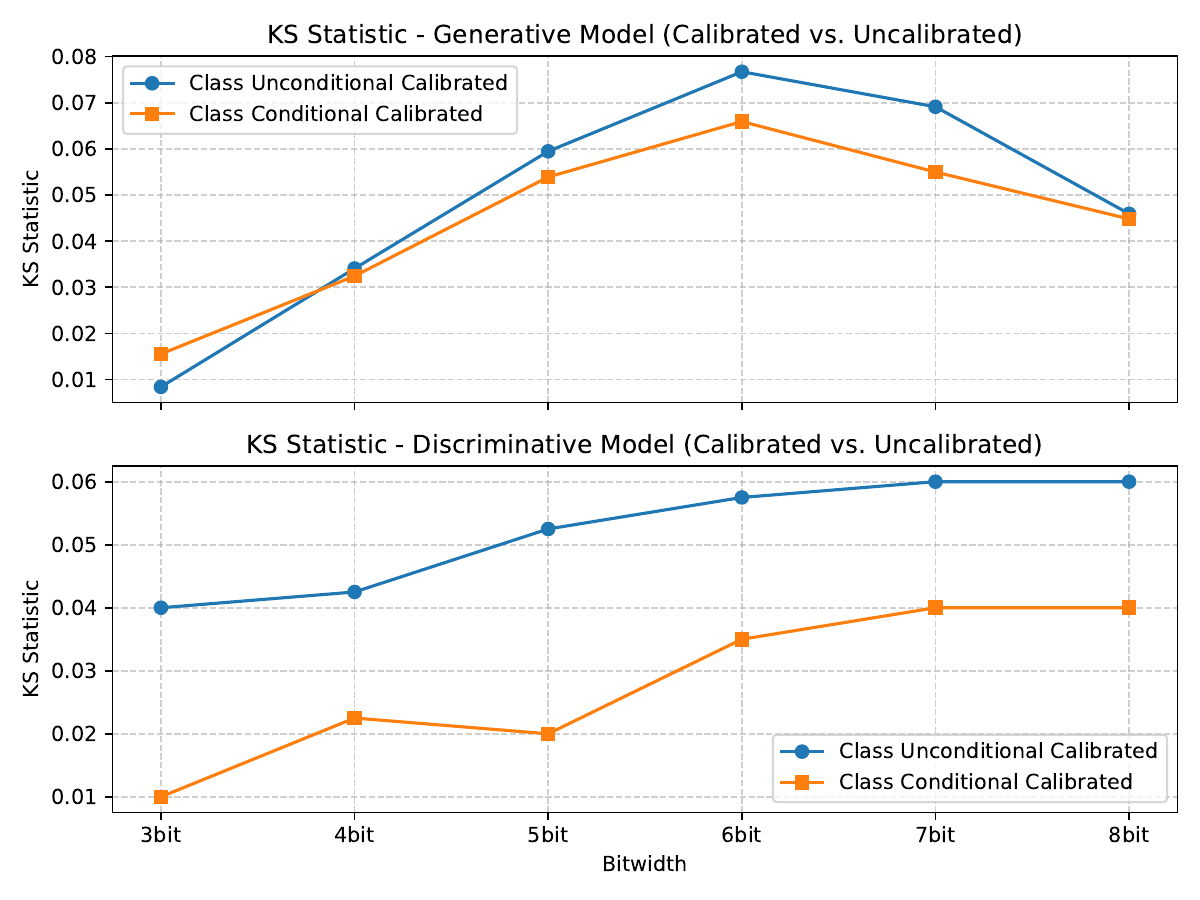}
    \caption{KS statistic comparison of weight shifts in generative and discriminative classifiers after class-conditional vs. class-unconditional calibration and GPFQ.}
    \label{fig:weight_comparison}
\end{figure}

We quantify these shifts using the Kolmogorov–Smirnov (KS) statistic \cite{kolmogorov1933empirical, smirnov1948table}, computed between the weights of the linear layer after quantization and their full-precision counterparts. As shown in Figure~\ref{fig:weight_comparison}, the KS values for the class-unconditionally calibrated generative classifier are consistently higher than those of the class-conditional model, indicating more substantial deviation in weight space. This gap becomes particularly large at higher bit-widths (e.g., 6-bit and above), where the class-unconditional calibration leads to larger weight changes during GPFQ.

Interestingly, the KS gap narrows at lower bit-widths (e.g., 3-bit and 4-bit), which initially seems counterintuitive. However, this narrowing is associated with smaller adjustments conducted during GPFQ under class-unconditional calibration suggesting that the model fails to make the necessary parameter corrections to compensate for quantization-induced loss. As a result, the generative classifier calibrated with class-unconditional data fails to adapt adequately at low bit-widths, leading to a substantial drop in downstream accuracy.

In contrast, the discriminative classifier exhibits a different behavior. While the KS values are generally higher for class-unconditional calibration across all bit-widths, the gap between conditional and unconditional calibration remains approximately uniform. This stability in distributional shifts correlates with the relatively consistent performance of the discriminative classifier under both calibration schemes. The model is able to absorb quantization effects more evenly, maintaining better generalization even under limited class representation during calibration.

These findings reveal that the performance degradation observed in generative classifiers is directly tied to the interaction between calibration data composition and the model's ability to adjust parameters during GPFQ. In particular, the lack of class balance in calibration leads to non-representative activation statistics, which in turn results in insufficient weight adaptation, especially at low bit-widths.

\subsubsection{Impact of Calibration Data on Activation Distribution}\label{subsubsec:impact_of_cal_on_act}

To further understand the mechanism behind the weight divergence observed in class conditionally versus class unconditionally calibrated generative classifiers, we analyze the activation distributions at key layers of the network. The goal is to assess whether differences in calibration data composition result in distinct activation patterns, which could lead to diverging calibration and quantization behavior during GPFQ.

\begin{figure}[h]
    \centering
    \includegraphics[width=0.4\linewidth]{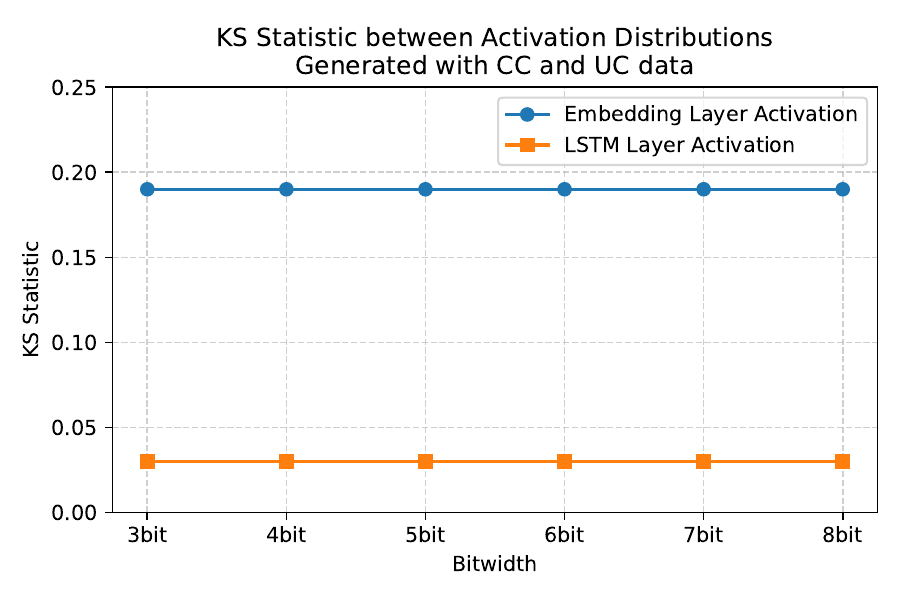}
    \caption{KS statistic between class-conditional (CC) and class-unconditional (UC) data activations in the embedding and LSTM layers of the pre-calibrated generative classifier.}
    \label{fig:activation_cc_uc}
\end{figure}

We first passed both class-conditional and class-unconditional calibration datasets through the same pre-calibrated generative classifier and extracted activation histograms from the embedding and LSTM layers. To quantify the statistical difference between these activation distributions, we apply the Kolmogorov–Smirnov (KS) test. The results are shown in Figure~\ref{fig:activation_cc_uc}.

The KS statistic values reveal a persistent and significant difference between the two calibration data types across all bit-widths. In the embedding layer, the KS value remains steady at approximately 0.19, while the LSTM layer shows a smaller but still consistent divergence around 0.03. These results confirm that class-unconditional and class-conditional data elicit meaningfully different activation behaviors and This discrepancy explains the differing behavior observed during GPFQ, since GPFQ adjusts model weights based on activation patterns.

\begin{figure}[h]
    \centering
    \includegraphics[width=0.4\linewidth]{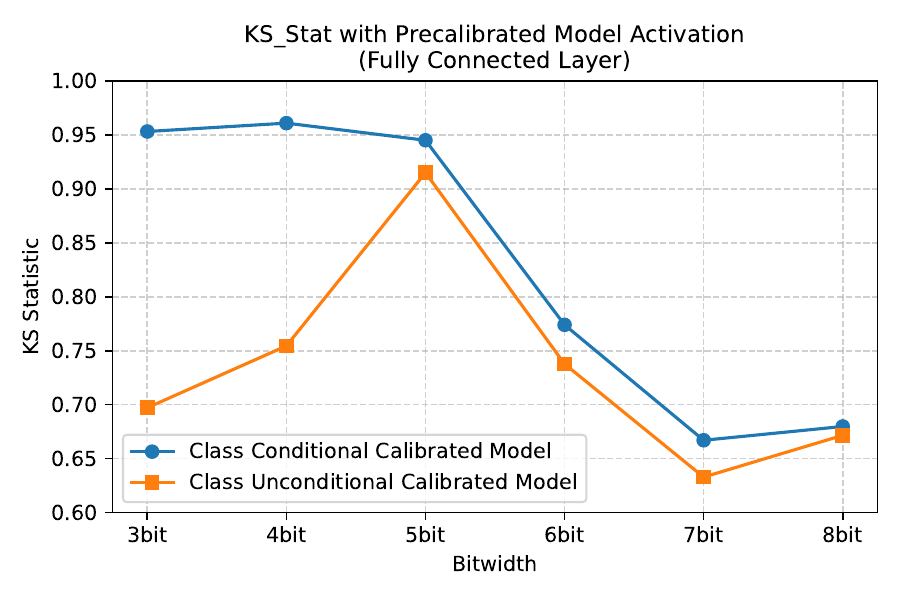}
    \caption{KS statistic between pre-calibrated model activations and class-conditional/class-unconditional calibrated model activations in the final linear layer across bit-widths.}
    \label{fig:activation_fc}
\end{figure}

To evaluate how these activation shifts affect the linear layer downstream, we conducted another experiment. We passed a common set of input sequences through three versions of the generative classifier: the pre-calibrated full-precision model, the class-conditional calibrated model, and the class-unconditional calibrated model. We then computed the KS statistic between the linear layer activations of each calibrated model and the pre-calibrated model, across various bit-widths. These results are presented in Figure~\ref{fig:activation_fc}.

The analysis shows that the class-conditional model exhibits higher KS values at 3-bit to 5-bit, indicating that larger corrective activation shifts occurred in the linear layer likely to compensate for quantization effects. In contrast, the class-unconditional calibrated model shows lower KS values in this range, suggesting insufficient adjustment. These insufficient shifts align with the earlier finding that weight adaptation under class-unconditional calibration is limited at low bit-widths, ultimately leading to worse performance.

Together, these findings point to a clear root cause: a mismatch in activation distributions between class-conditional and class-unconditional calibration datasets drives different internal representations, which propagate through the model and alter the effectiveness of GPFQ. This confirms that maintaining class balance in calibration data is critical for generative LSTM classifiers in order to ensure meaningful intermediate activations during quantization to prevent severe accuracy degradation.

\subsubsection{Loss Function Distribution Analysis}

To further understand the impact of quantization and input perturbations on the behavior of generative classifiers, we analyze the distribution of token-level cross-entropy losses across several configurations. Importantly, in generative classification, this cross-entropy loss serves as a proxy for the negative log-likelihood of the input sequence conditioned on the class label, as defined in Equation~\ref{eqn:likelihood}. Thus, deviations in token-level cross-entropy losses directly reflect distortions in the model’s estimate of the sequence likelihood, which forms the basis for accurate classification decisions.

We compare kernel density estimation (KDE) \cite{parzen1962, rosenblatt1956} plots of these token-level losses under four conditions: the full-precision model, the 3-bit quantized model calibrated with class-conditional data, the 3-bit model calibrated with class-unconditional data, and the 3-bit class-conditionally calibrated model evaluated on noisy input data.

\begin{figure*}[h]
    \centering
    \includegraphics[width=0.9\linewidth]{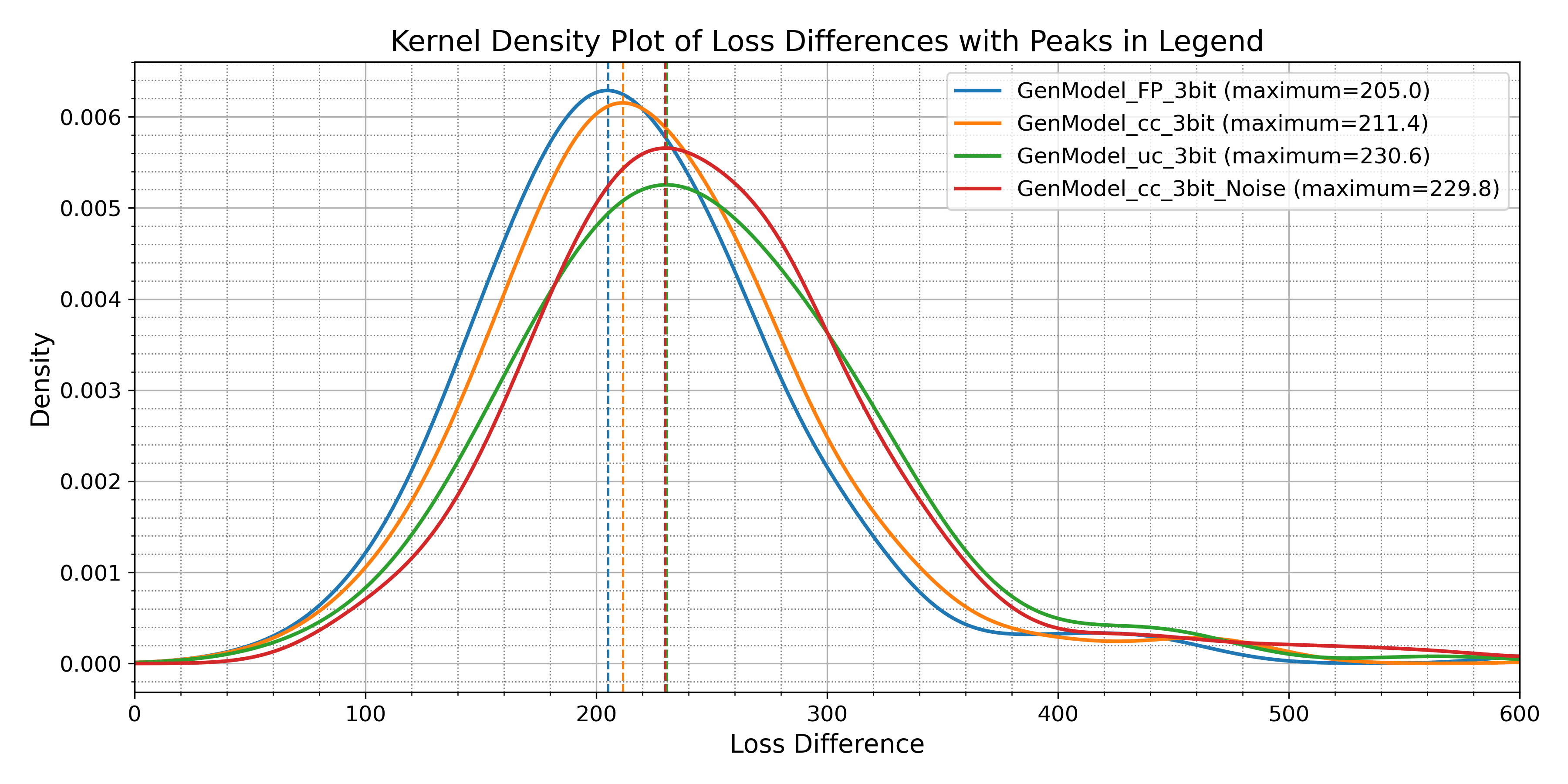}
    \caption{Kernel density plot showing loss difference (maximum values are shown in the legend) of token-level loss differences for the generative model (Full Precision, 3-bit class-conditional, 3-bit class-unconditional, and 3-bit class-conditional tested with noisy data). Rightward shifts indicate higher token-level loss in the class-unconditional and noisy-input settings.}
    \label{fig:gen_kde}
\end{figure*}

Figure~\ref{fig:gen_kde} illustrates the loss distributions for these settings. The 3-bit model calibrated with class-unconditional data shows a distinct rightward shift and broader tail, indicating that its predicted likelihoods are generally lower (i.e., higher cross entropy loss) compared to the full-precision and class-conditionally calibrated counterparts. This supports the performance drop observed in Section~\ref{subsec:PTQ_UC_Result} and suggests that class imbalance in calibration leads to misaligned or under-trained conditional likelihood estimates.

Additionally, when noise is injected during inference into the class-conditionally calibrated model, its loss distribution also shifts to the right, resembling that of the class-unconditional case. This degradation, discussed in Section~\ref{subsec:noise_test_Result}, highlights the compounding effect of input corruption in low bitwidth quantized models, even when calibration is class-balanced.

These KDE-based results reinforce that quantized generative models, particularly at low bitwidths, are highly sensitive to both calibration data diversity and input quality. Because these models base their classification decisions on likelihood maximization, distortions in token-level losses represent degraded confidence and sequence modeling accuracy, ultimately leading to sharp declines in classification performance.

\section{Conclusions and Future Work}

This work highlights a critical observation: the performance of generative text classifiers under PTQ is highly sensitive to class imbalance in the calibration dataset. We find that when PTQ is performed using unlabeled, class imbalanced data, the accuracy of quantized generative classifiers degrades significantly despite no changes to the structure of the network. This empirical result parallels earlier findings in the literature, where class imbalanced training data has been shown to negatively impact the representational capacity and distributional coverage of both discriminative and generative models, ultimately leading to a reduction in performance as evaluated using metrics such as precision and recall~\cite{kynkaanniemi2019precision,ward2021effect}. Our findings extend this understanding to the PTQ setting, suggesting that calibration of generative classifiers with imbalanced data can skew activation statistics and disproportionately favor majority-class representations, leading to unreliable  performance of quantized models.

We have presented a comprehensive comparative study of generative and discriminative LSTM-based text classification models under post-training quantization (PTQ) using the Brevitas framework, evaluating both model types across a range of bit-widths (3-bit to 8-bit) under various calibration strategies and noise conditions. Our findings reveal that generative classifiers exhibit strong robustness to input perturbations in full-precision settings and can maintain competitive performance under quantization when class-balanced calibration data is used. However, their performance degrades significantly under class-imbalanced calibration, where discriminative classifiers tend to outperform them, especially at lower bit-widths. Through controlled experiments, we demonstrate that class-conditional calibration combined with Greedy Path-Following Quantization (GPFQ) substantially improves generative model performance, while class-unconditional calibration fails to provide representative activation statistics, leading to insufficient weight adjustments and degraded accuracy. Kolmogorov–Smirnov (KS) tests further highlight distinct shifts in weight and activation distributions between class balanced and imbalanced calibration schemes, particularly for generative models. Furthermore, although generative classifiers are robust to noise in full-precision, this advantage diminishes after quantization particularly at low bit-widths underscoring a trade off between expressiveness and noise resilience under quantized deployment.

\begin{figure}[h]
    \centering
        \includegraphics[width=0.4\linewidth]{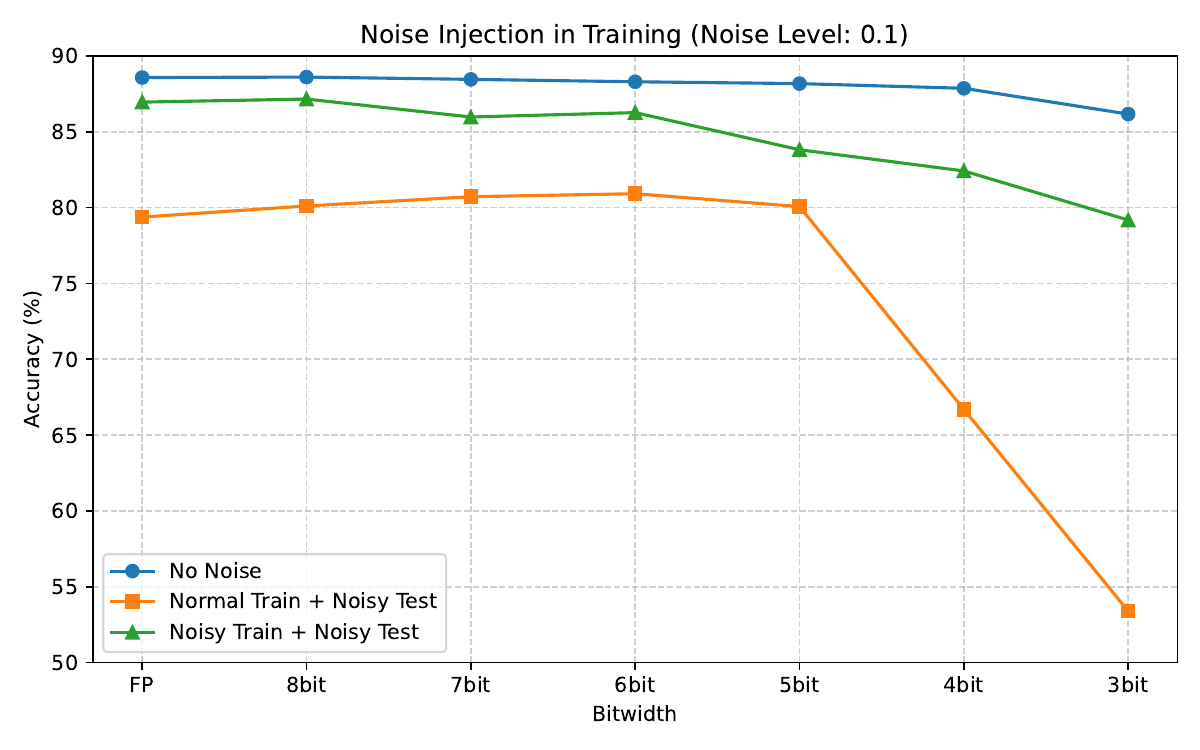}
    \caption{Impact of training-time noise injection on accuracy across bit-widths. Models are evaluated on noisy input with noise level 0.1. Training with noise improves robustness of quantized models at lower bit-widths.}
    \label{fig:noise_train}
\end{figure}

As a further exploration of robustness under quantization, we have conducted an experiment where noise was injected into the training data of the full-precision generative model. When this model is later quantized using class-conditional calibration and evaluated on similarly noisy input, we have observed improved resilience to accuracy degradation at lower bit-widths compared to the baseline model trained on clean data which has been shown in Figure~\ref{fig:noise_train}. This suggests that exposing the model to noise during training allows it to learn more stable representations, which carry over even after aggressive post-training quantization. This observation opens up future directions in combining robust training techniques with quantization workflows for better deployment reliability in noisy or error-prone environments.

Overall, this study offers the first systematic investigation into the post-training quantization performance of generative classifiers, especially in the context of LSTM-based text classification. Unlike the vast majority of prior work which focuses on quantizing discriminative models or generative models for synthesis, our findings reveal unique vulnerabilities of generative classifiers under PTQ particularly when calibration is performed on unlabeled, class-imbalanced data. Furthermore, our comparison with discriminative models highlights the differential sensitivity of the two paradigms to quantization and noise, providing important insights for real-world deployment. Finally, we emphasize the practical relevance of generative classifiers for edge computing due to their inherent robustness in full-precision settings, while also cautioning that this advantage can erode without proper calibration. These findings underscore the need for new PTQ strategies that account for model type, label availability, and calibration data quality to ensure reliable low-bit deployment in edge AI systems.

While our study focuses on LSTM-based models, similar investigations are needed for transformer-based text classifier architectures \cite{cunha2023comparative, sun2019fine}, which are increasingly used in edge NLP tasks. Future research should also explore hybrid calibration techniques that adaptively sample calibration data based on layer-wise quantization error profiles or activation distribution shifts. These metrics can help identify modules that are more prone to accuracy degradation under low-bit quantization. While Greedy Path-Following Quantization (GPFQ) has been effective for linear layers, its application to recurrent structures such as LSTM remains unexplored due to their internal gating and temporal dependencies. Extending GPFQ to quantize LSTM layers could open new opportunities for improving quantization fidelity in sequence based models such as LSTMs. In addition, techniques like mixed-precision quantization or layer-specific GPFQ could be explored to improve quantization outcomes for sensitive modules in generative classifiers. Finally, deploying these quantized models on actual edge devices with hardware-in-the-loop evaluation will provide further insight into their real-world efficiency and stability.

\section*{Acknowledgements}
\noindent The authors would like to acknowledge the Pacific Research Platform, NSF Project ACI-1541349, and Larry Smarr (PI, Calit2 at UCSD) for providing the computing infrastructure used in this project.

\bibliographystyle{cas-model2-names}

\bibliography{references}

\begin{thebibliography}{51}
\expandafter\ifx\csname natexlab\endcsname\relax\def\natexlab#1{#1}\fi
\providecommand{\url}[1]{\texttt{#1}}
\providecommand{\href}[2]{#2}
\providecommand{\path}[1]{#1}
\providecommand{\DOIprefix}{doi:}
\providecommand{\ArXivprefix}{arXiv:}
\providecommand{\URLprefix}{URL: }
\providecommand{\Pubmedprefix}{pmid:}
\providecommand{\doi}[1]{\href{http://dx.doi.org/#1}{\path{#1}}}
\providecommand{\Pubmed}[1]{\href{pmid:#1}{\path{#1}}}
\providecommand{\bibinfo}[2]{#2}
\ifx\xfnm\relax \def\xfnm[#1]{\unskip,\space#1}\fi
\bibitem[{Bishop(2006)}]{bishop2006pattern}
\bibinfo{author}{Bishop, C.M.}, \bibinfo{year}{2006}.
\newblock \bibinfo{title}{Pattern recognition and machine learning}. volume~\bibinfo{volume}{4}.
\newblock \bibinfo{publisher}{Springer}.
\bibitem[{Bondarenko et~al.(2021)Bondarenko, Nagel and Blankevoort}]{bondarenko2021understanding}
\bibinfo{author}{Bondarenko, Y.}, \bibinfo{author}{Nagel, M.}, \bibinfo{author}{Blankevoort, T.}, \bibinfo{year}{2021}.
\newblock \bibinfo{title}{Understanding and overcoming the challenges of efficient transformer quantization}.
\newblock \bibinfo{journal}{arXiv preprint arXiv:2109.12948} .
\bibitem[{Chiang et~al.(2023)Chiang, Li, Lin, Sheng, Wu, Zhang, Zheng, Zhuang, Zhuang, Gonzalez et~al.}]{chiang2023vicuna}
\bibinfo{author}{Chiang, W.L.}, \bibinfo{author}{Li, Z.}, \bibinfo{author}{Lin, Z.}, \bibinfo{author}{Sheng, Y.}, \bibinfo{author}{Wu, Z.}, \bibinfo{author}{Zhang, H.}, \bibinfo{author}{Zheng, L.}, \bibinfo{author}{Zhuang, S.}, \bibinfo{author}{Zhuang, Y.}, \bibinfo{author}{Gonzalez, J.E.}, et~al., \bibinfo{year}{2023}.
\newblock \bibinfo{title}{Vicuna: An open-source chatbot impressing gpt-4 with 90\%* chatgpt quality, march 2023}.
\newblock \bibinfo{journal}{URL https://lmsys. org/blog/2023-03-30-vicuna} \bibinfo{volume}{3}.
\bibitem[{Cunha et~al.(2023)Cunha, Viegas, Fran{\c{c}}a, Rosa, Rocha and Gon{\c{c}}alves}]{cunha2023comparative}
\bibinfo{author}{Cunha, W.}, \bibinfo{author}{Viegas, F.}, \bibinfo{author}{Fran{\c{c}}a, C.}, \bibinfo{author}{Rosa, T.}, \bibinfo{author}{Rocha, L.}, \bibinfo{author}{Gon{\c{c}}alves, M.A.}, \bibinfo{year}{2023}.
\newblock \bibinfo{title}{A comparative survey of instance selection methods applied to non-neural and transformer-based text classification}.
\newblock \bibinfo{journal}{ACM Computing Surveys} \bibinfo{volume}{55}, \bibinfo{pages}{1--52}.
\bibitem[{Ding and Gimpel(2019)}]{ding2019latent}
\bibinfo{author}{Ding, X.}, \bibinfo{author}{Gimpel, K.}, \bibinfo{year}{2019}.
\newblock \bibinfo{title}{Latent-variable generative models for data-efficient text classification}.
\newblock \bibinfo{journal}{arXiv preprint arXiv:1910.00382} \URLprefix \url{https://arxiv.org/abs/1910.00382}, \href{http://arxiv.org/abs/1910.00382}{\tt arXiv:1910.00382}.
\bibitem[{Eamaz et~al.(2023)Eamaz, Yeganegi and Soltanalian}]{eamaz2023matrix}
\bibinfo{author}{Eamaz, A.}, \bibinfo{author}{Yeganegi, F.}, \bibinfo{author}{Soltanalian, M.}, \bibinfo{year}{2023}.
\newblock \bibinfo{title}{Matrix completion via memoryless scalar quantization}.
\newblock \bibinfo{journal}{arXiv preprint arXiv:2311.05052} .
\bibitem[{Esser et~al.(2019)Esser, McKinstry, Bablani, Appuswamy and Modha}]{esser2019learned}
\bibinfo{author}{Esser, S.K.}, \bibinfo{author}{McKinstry, J.L.}, \bibinfo{author}{Bablani, D.}, \bibinfo{author}{Appuswamy, R.}, \bibinfo{author}{Modha, D.S.}, \bibinfo{year}{2019}.
\newblock \bibinfo{title}{Learned step size quantization}.
\newblock \bibinfo{journal}{arXiv preprint arXiv:1902.08153} .
\bibitem[{Franco et~al.(2025)Franco, Pappalardo and Fraser}]{brevitas}
\bibinfo{author}{Franco, G.}, \bibinfo{author}{Pappalardo, A.}, \bibinfo{author}{Fraser, N.J.}, \bibinfo{year}{2025}.
\newblock \bibinfo{title}{Xilinx/brevitas}.
\newblock \URLprefix \url{https://doi.org/10.5281/zenodo.3333552}, \DOIprefix\doi{10.5281/zenodo.3333552}.
\bibitem[{Hastie et~al.(2009)Hastie, Tibshirani, Friedman and Friedman}]{hastie2009elements}
\bibinfo{author}{Hastie, T.}, \bibinfo{author}{Tibshirani, R.}, \bibinfo{author}{Friedman, J.H.}, \bibinfo{author}{Friedman, J.H.}, \bibinfo{year}{2009}.
\newblock \bibinfo{title}{The elements of statistical learning: data mining, inference, and prediction}. volume~\bibinfo{volume}{2}.
\newblock \bibinfo{publisher}{Springer}.
\bibitem[{Hochreiter and Schmidhuber(1997)}]{hochreiter1997long}
\bibinfo{author}{Hochreiter, S.}, \bibinfo{author}{Schmidhuber, J.}, \bibinfo{year}{1997}.
\newblock \bibinfo{title}{Long short-term memory}.
\newblock \bibinfo{journal}{Neural computation} \bibinfo{volume}{9}, \bibinfo{pages}{1735--1780}.
\bibitem[{Honnibal et~al.(2020)Honnibal, Montani, Van~Landeghem, Boyd et~al.}]{honnibal2020spacy}
\bibinfo{author}{Honnibal, M.}, \bibinfo{author}{Montani, I.}, \bibinfo{author}{Van~Landeghem, S.}, \bibinfo{author}{Boyd, A.}, et~al., \bibinfo{year}{2020}.
\newblock \bibinfo{title}{spacy: Industrial-strength natural language processing in python} .
\bibitem[{Howard and Gugger(2020)}]{howard2020}
\bibinfo{author}{Howard, J.}, \bibinfo{author}{Gugger, S.}, \bibinfo{year}{2020}.
\newblock \bibinfo{title}{Fastai: A layered api for deep learning}.
\newblock \bibinfo{journal}{Information} \bibinfo{volume}{11}, \bibinfo{pages}{108}.
\bibitem[{Hu et~al.(2022)Hu, Meinel and Yang}]{hu2022empirical}
\bibinfo{author}{Hu, T.}, \bibinfo{author}{Meinel, C.}, \bibinfo{author}{Yang, H.}, \bibinfo{year}{2022}.
\newblock \bibinfo{title}{Empirical evaluation of post-training quantization methods for language tasks}.
\newblock \bibinfo{journal}{arXiv preprint arXiv:2210.16621} .
\bibitem[{Hubara et~al.(2021)Hubara, Nahshan, Hanani, Banner and Soudry}]{hubara2021accurate}
\bibinfo{author}{Hubara, I.}, \bibinfo{author}{Nahshan, Y.}, \bibinfo{author}{Hanani, Y.}, \bibinfo{author}{Banner, R.}, \bibinfo{author}{Soudry, D.}, \bibinfo{year}{2021}.
\newblock \bibinfo{title}{Accurate post training quantization with small calibration sets}, in: \bibinfo{booktitle}{International Conference on Machine Learning}, \bibinfo{organization}{PMLR}. pp. \bibinfo{pages}{4466--4475}.
\bibitem[{Jurafsky and Martin(2023)}]{jurafsky2023speech}
\bibinfo{author}{Jurafsky, D.}, \bibinfo{author}{Martin, J.H.}, \bibinfo{year}{2023}.
\newblock \bibinfo{title}{Speech and Language Processing}.
\newblock \bibinfo{edition}{3rd} ed., \bibinfo{publisher}{Pearson}.
\newblock \bibinfo{note}{Draft version available at \url{https://web.stanford.edu/~jurafsky/slp3/}}.
\bibitem[{Kolmogorov(1933)}]{kolmogorov1933empirical}
\bibinfo{author}{Kolmogorov, A.N.}, \bibinfo{year}{1933}.
\newblock \bibinfo{title}{Sulla determinazione empirica di una legge di distribuzione}.
\newblock \bibinfo{journal}{Giornale dell'Istituto Italiano degli Attuari} \bibinfo{volume}{4}, \bibinfo{pages}{83--91}.
\bibitem[{Ktena et~al.(2024)Ktena, Wiles, Albuquerque, Rebuffi, Tanno, Roy, Azizi, Belgrave, Kohli, Cemgil et~al.}]{ktena2024generative}
\bibinfo{author}{Ktena, I.}, \bibinfo{author}{Wiles, O.}, \bibinfo{author}{Albuquerque, I.}, \bibinfo{author}{Rebuffi, S.A.}, \bibinfo{author}{Tanno, R.}, \bibinfo{author}{Roy, A.G.}, \bibinfo{author}{Azizi, S.}, \bibinfo{author}{Belgrave, D.}, \bibinfo{author}{Kohli, P.}, \bibinfo{author}{Cemgil, T.}, et~al., \bibinfo{year}{2024}.
\newblock \bibinfo{title}{Generative models improve fairness of medical classifiers under distribution shifts}.
\newblock \bibinfo{journal}{Nature Medicine} \bibinfo{volume}{30}, \bibinfo{pages}{1166--1173}.
\bibitem[{Kynk{\"a}{\"a}nniemi et~al.(2019)Kynk{\"a}{\"a}nniemi, Karras, Laine, Lehtinen and Aila}]{kynkaanniemi2019precision}
\bibinfo{author}{Kynk{\"a}{\"a}nniemi, T.}, \bibinfo{author}{Karras, T.}, \bibinfo{author}{Laine, S.}, \bibinfo{author}{Lehtinen, J.}, \bibinfo{author}{Aila, T.}, \bibinfo{year}{2019}.
\newblock \bibinfo{title}{Precision and recall metrics for generative models}, in: \bibinfo{booktitle}{Advances in Neural Information Processing Systems}.
\newblock \URLprefix \url{https://proceedings.neurips.cc/paper_files/paper/2018/file/f7696a9b362ac5a51c3dc8f098b73923-Paper.pdf}.
\bibitem[{Lee et~al.(2019)Lee, Yun, Lee, Lee, Li and Shin}]{lee2019robust}
\bibinfo{author}{Lee, K.}, \bibinfo{author}{Yun, S.}, \bibinfo{author}{Lee, K.}, \bibinfo{author}{Lee, H.}, \bibinfo{author}{Li, B.}, \bibinfo{author}{Shin, J.}, \bibinfo{year}{2019}.
\newblock \bibinfo{title}{Robust inference via generative classifiers for handling noisy labels}, in: \bibinfo{booktitle}{Proceedings of the 36th International Conference on Machine Learning (ICML)}.
\bibitem[{Li et~al.(2023a)Li, Zhang, Yen and Xu}]{li2023fp8}
\bibinfo{author}{Li, J.}, \bibinfo{author}{Zhang, T.}, \bibinfo{author}{Yen, I.E.H.}, \bibinfo{author}{Xu, D.}, \bibinfo{year}{2023}a.
\newblock \bibinfo{title}{Fp8-bert: Post-training quantization for transformer}.
\newblock \bibinfo{journal}{arXiv preprint arXiv:2312.05725} .
\bibitem[{Li et~al.(2023b)Li, Ganesh, Yao, Jain, Gholami and Gonzalez}]{li2023impact}
\bibinfo{author}{Li, S.}, \bibinfo{author}{Ganesh, V.}, \bibinfo{author}{Yao, Z.}, \bibinfo{author}{Jain, P.}, \bibinfo{author}{Gholami, A.}, \bibinfo{author}{Gonzalez, J.E.}, \bibinfo{year}{2023}b.
\newblock \bibinfo{title}{On the impact of calibration data in post-training quantization and pruning}.
\newblock \bibinfo{journal}{arXiv preprint arXiv:2311.09755} .
\bibitem[{Li et~al.(2018)Li, Bradshaw and Sharma}]{li2018generative_robustness}
\bibinfo{author}{Li, Y.}, \bibinfo{author}{Bradshaw, J.}, \bibinfo{author}{Sharma, Y.}, \bibinfo{year}{2018}.
\newblock \bibinfo{title}{Are generative classifiers more robust to adversarial attacks?}
\newblock \bibinfo{journal}{arXiv preprint arXiv:1802.06552} .
\bibitem[{Liu et~al.(2024)Liu, Zhao, Fedorov, Soran, Choudhary, Krishnamoorthi, Chandra, Tian and Blankevoort}]{liu2024spinquant}
\bibinfo{author}{Liu, Z.}, \bibinfo{author}{Zhao, C.}, \bibinfo{author}{Fedorov, I.}, \bibinfo{author}{Soran, B.}, \bibinfo{author}{Choudhary, D.}, \bibinfo{author}{Krishnamoorthi, R.}, \bibinfo{author}{Chandra, V.}, \bibinfo{author}{Tian, Y.}, \bibinfo{author}{Blankevoort, T.}, \bibinfo{year}{2024}.
\newblock \bibinfo{title}{Spinquant: Llm quantization with learned rotations}.
\newblock \bibinfo{journal}{arXiv preprint arXiv:2405.16406} .
\bibitem[{Lybrand and Saab(2021)}]{lybrand2021greedy}
\bibinfo{author}{Lybrand, E.}, \bibinfo{author}{Saab, R.}, \bibinfo{year}{2021}.
\newblock \bibinfo{title}{A greedy algorithm for quantizing neural networks}.
\newblock \bibinfo{journal}{Journal of Machine Learning Research} \bibinfo{volume}{22}, \bibinfo{pages}{1--38}.
\newblock \URLprefix \url{http://jmlr.org/papers/v22/20-1233.html}. \bibinfo{note}{arXiv:2010.15979}.
\bibitem[{Mikolov et~al.(2013)Mikolov, Chen, Corrado and Dean}]{mikolov2013efficient}
\bibinfo{author}{Mikolov, T.}, \bibinfo{author}{Chen, K.}, \bibinfo{author}{Corrado, G.}, \bibinfo{author}{Dean, J.}, \bibinfo{year}{2013}.
\newblock \bibinfo{title}{Efficient estimation of word representations in vector space}.
\newblock \bibinfo{journal}{arXiv preprint arXiv:1301.3781} .
\bibitem[{Nagel et~al.(2020)Nagel, Amjad, van Baalen, Louizos and Blankevoort}]{nagel2020up}
\bibinfo{author}{Nagel, M.}, \bibinfo{author}{Amjad, R.A.}, \bibinfo{author}{van Baalen, M.}, \bibinfo{author}{Louizos, C.}, \bibinfo{author}{Blankevoort, T.}, \bibinfo{year}{2020}.
\newblock \bibinfo{title}{Up or down? adaptive rounding for post-training quantization}.
\newblock \bibinfo{journal}{CoRR} \bibinfo{volume}{abs/2004.10568}.
\newblock \URLprefix \url{https://arxiv.org/abs/2004.10568}, \href{http://arxiv.org/abs/2004.10568}{\tt arXiv:2004.10568}.
\bibitem[{Nagel et~al.(2019)Nagel, Baalen, Blankevoort and Welling}]{nagel2019data}
\bibinfo{author}{Nagel, M.}, \bibinfo{author}{Baalen, M.v.}, \bibinfo{author}{Blankevoort, T.}, \bibinfo{author}{Welling, M.}, \bibinfo{year}{2019}.
\newblock \bibinfo{title}{Data-free quantization through weight equalization and bias correction}, in: \bibinfo{booktitle}{Proceedings of the IEEE/CVF international conference on computer vision}, pp. \bibinfo{pages}{1325--1334}.
\bibitem[{Nagel et~al.(2021)}]{nagel2021}
\bibinfo{author}{Nagel, M.}, et~al., \bibinfo{year}{2021}.
\newblock \bibinfo{title}{A white paper on neural network quantization}.
\newblock \bibinfo{journal}{arXiv preprint arXiv:2106.08295} .
\bibitem[{Ng and Jordan(2001)}]{ng2001discriminative}
\bibinfo{author}{Ng, A.}, \bibinfo{author}{Jordan, M.}, \bibinfo{year}{2001}.
\newblock \bibinfo{title}{On discriminative vs. generative classifiers: A comparison of logistic regression and naive bayes}.
\newblock \bibinfo{journal}{Advances in neural information processing systems} \bibinfo{volume}{14}.
\bibitem[{Nguyen et~al.(2015)Nguyen, Yosinski and Clune}]{nguyen2015deep}
\bibinfo{author}{Nguyen, A.}, \bibinfo{author}{Yosinski, J.}, \bibinfo{author}{Clune, J.}, \bibinfo{year}{2015}.
\newblock \bibinfo{title}{Deep neural networks are easily fooled: High confidence predictions for unrecognizable images}, in: \bibinfo{booktitle}{Proceedings of the IEEE conference on computer vision and pattern recognition}, pp. \bibinfo{pages}{427--436}.
\bibitem[{Parzen(1962)}]{parzen1962}
\bibinfo{author}{Parzen, E.}, \bibinfo{year}{1962}.
\newblock \bibinfo{title}{On estimation of a probability density function and mode}.
\newblock \bibinfo{journal}{The Annals of Mathematical Statistics} \bibinfo{volume}{33}, \bibinfo{pages}{1065--1076}.
\bibitem[{Paszke(2019)}]{paszke2019pytorch}
\bibinfo{author}{Paszke, A.}, \bibinfo{year}{2019}.
\newblock \bibinfo{title}{Pytorch: An imperative style, high-performance deep learning library}.
\newblock \bibinfo{journal}{arXiv preprint arXiv:1912.01703} .
\bibitem[{Pennington et~al.(2014)Pennington, Socher and Manning}]{pennington2014glove}
\bibinfo{author}{Pennington, J.}, \bibinfo{author}{Socher, R.}, \bibinfo{author}{Manning, C.D.}, \bibinfo{year}{2014}.
\newblock \bibinfo{title}{{GloVe}: Global vectors for word representation}, in: \bibinfo{booktitle}{Proceedings of the 2014 Conference on Empirical Methods in Natural Language Processing (EMNLP)}, pp. \bibinfo{pages}{1532--1543}.
\bibitem[{Rosenblatt(1956)}]{rosenblatt1956}
\bibinfo{author}{Rosenblatt, M.}, \bibinfo{year}{1956}.
\newblock \bibinfo{title}{Remarks on some nonparametric estimates of a density function}.
\newblock \bibinfo{journal}{The Annals of Mathematical Statistics} \bibinfo{volume}{27}, \bibinfo{pages}{832--837}.
\bibitem[{Ruan et~al.(2021)Ruan, Colbert, Kreutz-Delgado and Das}]{ruan2021generative}
\bibinfo{author}{Ruan, S.}, \bibinfo{author}{Colbert, I.}, \bibinfo{author}{Kreutz-Delgado, K.}, \bibinfo{author}{Das, S.}, \bibinfo{year}{2021}.
\newblock \bibinfo{title}{Generative and discriminative deep belief network classifiers: Comparisons under an approximate computing framework}.
\newblock \bibinfo{journal}{arXiv preprint arXiv:2102.00534} .
\bibitem[{Smirnov(1948)}]{smirnov1948table}
\bibinfo{author}{Smirnov, N.}, \bibinfo{year}{1948}.
\newblock \bibinfo{title}{Table for estimating the goodness of fit of empirical distributions}.
\newblock \bibinfo{journal}{Annals of Mathematical Statistics} \bibinfo{volume}{19}, \bibinfo{pages}{279--281}.
\bibitem[{Sun et~al.(2019)Sun, Qiu, Xu and Huang}]{sun2019fine}
\bibinfo{author}{Sun, C.}, \bibinfo{author}{Qiu, X.}, \bibinfo{author}{Xu, Y.}, \bibinfo{author}{Huang, X.}, \bibinfo{year}{2019}.
\newblock \bibinfo{title}{How to fine-tune bert for text classification?}, in: \bibinfo{booktitle}{China national conference on Chinese computational linguistics}, \bibinfo{organization}{Springer}. pp. \bibinfo{pages}{194--206}.
\bibitem[{Touvron et~al.(2023)Touvron, Lavril, Izacard, Martinet, Lachaux, Lacroix, Rozi{\`e}re, Goyal, Hambro, Azhar et~al.}]{touvron2023llama}
\bibinfo{author}{Touvron, H.}, \bibinfo{author}{Lavril, T.}, \bibinfo{author}{Izacard, G.}, \bibinfo{author}{Martinet, X.}, \bibinfo{author}{Lachaux, M.A.}, \bibinfo{author}{Lacroix, T.}, \bibinfo{author}{Rozi{\`e}re, B.}, \bibinfo{author}{Goyal, N.}, \bibinfo{author}{Hambro, E.}, \bibinfo{author}{Azhar, F.}, et~al., \bibinfo{year}{2023}.
\newblock \bibinfo{title}{Llama: Open and efficient foundation language models}.
\newblock \bibinfo{journal}{arXiv preprint arXiv:2302.13971} .
\bibitem[{Vaswani et~al.(2017)Vaswani, Shazeer, Parmar, Uszkoreit, Jones, Gomez, Kaiser and Polosukhin}]{vaswani2017attention}
\bibinfo{author}{Vaswani, A.}, \bibinfo{author}{Shazeer, N.}, \bibinfo{author}{Parmar, N.}, \bibinfo{author}{Uszkoreit, J.}, \bibinfo{author}{Jones, L.}, \bibinfo{author}{Gomez, A.N.}, \bibinfo{author}{Kaiser, {\L}.}, \bibinfo{author}{Polosukhin, I.}, \bibinfo{year}{2017}.
\newblock \bibinfo{title}{Attention is all you need}.
\newblock \bibinfo{journal}{Advances in neural information processing systems} \bibinfo{volume}{30}.
\bibitem[{Wang et~al.(2019)Wang, Liu, Wu, Yang and Han}]{wang2019haq}
\bibinfo{author}{Wang, K.}, \bibinfo{author}{Liu, Z.}, \bibinfo{author}{Wu, Y.}, \bibinfo{author}{Yang, J.}, \bibinfo{author}{Han, S.}, \bibinfo{year}{2019}.
\newblock \bibinfo{title}{Haq: Hardware-aware automated quantization with mixed precision}.
\newblock \bibinfo{journal}{Proceedings of the IEEE/CVF Conference on Computer Vision and Pattern Recognition (CVPR)} , \bibinfo{pages}{8612--8620}\href{http://arxiv.org/abs/1811.08886}{\tt arXiv:1811.08886}.
\bibitem[{Ward and Williams(2021)}]{ward2021effect}
\bibinfo{author}{Ward, R.}, \bibinfo{author}{Williams, C.K.}, \bibinfo{year}{2021}.
\newblock \bibinfo{title}{The effect of class imbalance on precision-recall curves}.
\newblock \bibinfo{journal}{Neural Computation} \bibinfo{volume}{33}, \bibinfo{pages}{853--889}.
\newblock \URLprefix \url{https://direct.mit.edu/neco/article/33/4/853/97475/The-Effect-of-Class-Imbalance-on-Precision-Recall}, \DOIprefix\doi{10.1162/neco_a_01370}.
\bibitem[{Xu et~al.(2022)Xu, Yuan, Wu, Yu and Wang}]{xu2022easyquant}
\bibinfo{author}{Xu, C.}, \bibinfo{author}{Yuan, L.}, \bibinfo{author}{Wu, Y.}, \bibinfo{author}{Yu, W.}, \bibinfo{author}{Wang, L.}, \bibinfo{year}{2022}.
\newblock \bibinfo{title}{Easyquant: Post-training quantization via scale optimization}, in: \bibinfo{booktitle}{CVPR}.
\bibitem[{Yao et~al.(2022)Yao, Yazdani~Aminabadi, Zhang, Wu, Li and He}]{yao2022zeroquant}
\bibinfo{author}{Yao, Z.}, \bibinfo{author}{Yazdani~Aminabadi, R.}, \bibinfo{author}{Zhang, M.}, \bibinfo{author}{Wu, X.}, \bibinfo{author}{Li, C.}, \bibinfo{author}{He, Y.}, \bibinfo{year}{2022}.
\newblock \bibinfo{title}{Zeroquant: Efficient and affordable post-training quantization for large-scale transformers}.
\newblock \bibinfo{journal}{Advances in Neural Information Processing Systems} \bibinfo{volume}{35}, \bibinfo{pages}{27168--27183}.
\bibitem[{Yogatama et~al.(2017)Yogatama, Dyer, Ling and Blunsom}]{yogatama2017generative}
\bibinfo{author}{Yogatama, D.}, \bibinfo{author}{Dyer, C.}, \bibinfo{author}{Ling, W.}, \bibinfo{author}{Blunsom, P.}, \bibinfo{year}{2017}.
\newblock \bibinfo{title}{Generative and discriminative text classification with recurrent neural networks}.
\newblock \bibinfo{journal}{arXiv preprint arXiv:1703.01898} \URLprefix \url{https://arxiv.org/abs/1703.01898}, \href{http://arxiv.org/abs/1703.01898}{\tt arXiv:1703.01898}.
\bibitem[{Zafrir et~al.(2019)Zafrir, Boudoukh, Izsak and Wasserblat}]{zafrir2019q8bert}
\bibinfo{author}{Zafrir, O.}, \bibinfo{author}{Boudoukh, G.}, \bibinfo{author}{Izsak, P.}, \bibinfo{author}{Wasserblat, M.}, \bibinfo{year}{2019}.
\newblock \bibinfo{title}{Q8bert: Quantized 8bit bert}.
\newblock \bibinfo{journal}{arXiv preprint arXiv:1910.06188} .
\bibitem[{Zhang et~al.(2022a)Zhang, Roller, Goyal, Artetxe et~al.}]{zhang2022opt}
\bibinfo{author}{Zhang, S.}, \bibinfo{author}{Roller, S.}, \bibinfo{author}{Goyal, N.}, \bibinfo{author}{Artetxe, M.}, et~al., \bibinfo{year}{2022}a.
\newblock \bibinfo{title}{Opt: Open pre-trained transformer language models}.
\newblock \bibinfo{journal}{arXiv preprint arXiv:2205.01068} .
\bibitem[{Zhang et~al.(2025a)Zhang, Zhang, Colbert and Saab}]{zhang2025qronos}
\bibinfo{author}{Zhang, S.}, \bibinfo{author}{Zhang, H.}, \bibinfo{author}{Colbert, I.}, \bibinfo{author}{Saab, R.}, \bibinfo{year}{2025}a.
\newblock \bibinfo{title}{Qronos: Correcting the past by shaping the future... in post-training quantization}.
\newblock \bibinfo{journal}{arXiv preprint arXiv:2505.11695} .
\bibitem[{Zhang et~al.(2022b)Zhang, Colbert and Das}]{zhang2022learning}
\bibinfo{author}{Zhang, X.}, \bibinfo{author}{Colbert, I.}, \bibinfo{author}{Das, S.}, \bibinfo{year}{2022}b.
\newblock \bibinfo{title}{Learning low-precision structured subnetworks using joint layerwise channel pruning and uniform quantization}.
\newblock \bibinfo{journal}{Applied Sciences} \bibinfo{volume}{12}, \bibinfo{pages}{7829}.
\bibitem[{Zhang et~al.(2015)Zhang, Zhao and LeCun}]{zhang2015_AgNews_DBPedia}
\bibinfo{author}{Zhang, X.}, \bibinfo{author}{Zhao, J.}, \bibinfo{author}{LeCun, Y.}, \bibinfo{year}{2015}.
\newblock \bibinfo{title}{Character-level convolutional networks for text classification}, in: \bibinfo{booktitle}{Advances in Neural Information Processing Systems}.
\newblock \URLprefix \url{https://arxiv.org/abs/1509.01626}.
\bibitem[{Zhang and Wallace(2015)}]{zhang2015}
\bibinfo{author}{Zhang, Y.}, \bibinfo{author}{Wallace, B.C.}, \bibinfo{year}{2015}.
\newblock \bibinfo{title}{A sensitivity analysis of (and practitioners' guide to) convolutional neural networks for sentence classification}.
\newblock \bibinfo{journal}{arXiv preprint arXiv:1510.03820} .
\bibitem[{Zhang et~al.(2025b)Zhang, Gao, Fan, Zhao, Yang and Yan}]{zhang2025selectq}
\bibinfo{author}{Zhang, Z.}, \bibinfo{author}{Gao, Y.}, \bibinfo{author}{Fan, J.}, \bibinfo{author}{Zhao, Z.}, \bibinfo{author}{Yang, Y.}, \bibinfo{author}{Yan, S.}, \bibinfo{year}{2025}b.
\newblock \bibinfo{title}{Selectq: Calibration data selection for post-training quantization}.
\newblock \bibinfo{journal}{Machine Intelligence Research} , \bibinfo{pages}{1--12}.

\end{thebibliography}





\end{document}